\documentclass[sigconf]{acmart}

\AtBeginDocument{\providecommand\BibTeX{{\normalfont B\kern-0.5em{\scshape i\kern-0.25em b}\kern-0.8em\TeX}}}

\setcopyright{acmcopyright}

\acmYear{2021}
\copyrightyear{2021}
\setcopyright{acmlicensed}
\acmConference[MobiSys '21]{The 19th Annual International Conference on Mobile Systems, Applications, and Services}{June 24--July 2, 2021}{Virtual, WI, USA}
\acmBooktitle{The 19th Annual International Conference on Mobile Systems, Applications, and Services (MobiSys '21), June 24--July 2, 2021, Virtual, WI, USA}
\acmPrice{15.00}
\acmDOI{10.1145/3458864.3467886}
\acmISBN{978-1-4503-8443-8/21/06}

\usepackage{xspace}
\usepackage{balance}
\usepackage{xcolor,color}
\usepackage{amsmath}
\usepackage[textfont={normal,bf, it},labelfont=bf]{caption}
\usepackage[labelformat=simple,textfont=small]{subcaption}

\usepackage{breakurl}
\PassOptionsToPackage{hyphens}{url}\usepackage{hyperref}

\usepackage{enumitem}
\usepackage{booktabs}

\newcommand{\ra}[1]{\renewcommand{\arraystretch}{#1}}

\newcommand{\sysname}{\textsc{Xihe}\xspace}

\newcommand{\SHc}{SH coefficients\xspace}

\newcommand{\uspcs}{unit sphere-based point cloud sampling\xspace}
\newcommand{\uspc}{unit sphere-based point cloud\xspace}

\newcommand{\svl}{spatially-variant lighting\xspace}

\newcommand{\xiheNet}{\textsc{XiheNet}\xspace}

\definecolor{pro_green}{rgb}{0.0, 0.66, 0.47}
\definecolor{overleaf_green}{rgb}{0.08, 0.54, 0.02}

\newcommand{\1}{{\em (i)}}
\newcommand{\2}{{\em (ii)}}
\newcommand{\3}{{\em (iii)}}

\newcommand{\para }[1]{\noindent  {\bf \emph{#1}}}

\begin{document}

\title{Xihe: A 3D Vision-based Lighting Estimation Framework for Mobile Augmented Reality}

\author{Yiqin Zhao}
\affiliation{\institution{Worcester Polytechnic Institute}
 \streetaddress{100 Institute Road}
 \city{}
 \state{}
 \country{}
 }
\email{yzhao11@wpi.edu}

\author{Tian Guo}
\affiliation{\institution{Worcester Polytechnic Institute}
 \streetaddress{100 Institute Road}
 \city{}
 \state{}
 \country{}
 }
\email{tian@wpi.edu}

\renewcommand{\shortauthors}{Y. Zhao and T. Guo}

\begin{abstract}
Omnidirectional lighting provides the foundation for achieving spatially-variant photorealistic 3D rendering, a desirable property for mobile augmented reality applications.
However, in practice, estimating omnidirectional lighting can be challenging due to limitations such as partial panoramas of the rendering positions, and the inherent environment lighting and mobile user dynamics.
A new opportunity arises recently with the advancements in mobile 3D vision, including built-in high-accuracy depth sensors and deep learning-powered algorithms, which provide the means to better sense and understand the physical surroundings.
Centering the key idea of 3D vision, in this work, we design
an edge-assisted framework called \sysname to provide
mobile AR applications the ability to obtain accurate omnidirectional lighting estimation in real time.

Specifically, we develop a novel sampling technique that efficiently compresses the raw point cloud input generated at the mobile device. This technique is derived based on our empirical analysis of a recent 3D indoor dataset and plays a key role in our 3D vision-based lighting estimator pipeline design.
To achieve the real-time goal, we develop a tailored GPU pipeline for on-device point cloud processing and use an encoding technique that reduces network transmitted bytes.
Finally, we present an adaptive triggering strategy that allows \sysname to skip unnecessary lighting estimations and a practical way to provide temporal coherent rendering integration with the mobile AR ecosystem.
We evaluate both the lighting estimation accuracy and time of \sysname using a reference mobile application developed with \sysname's APIs. Our results show that \sysname takes as fast as 20.67ms per lighting estimation and achieves 9.4\% better estimation accuracy than a state-of-the-art neural network.
\end{abstract}



\begin{abstract}
Omnidirectional lighting provides the foundation for achieving spatially-variant photorealistic 3D rendering, a desirable property for mobile augmented reality applications.
However, in practice, estimating omnidirectional lighting can be challenging due to limitations such as partial panoramas of the rendering positions, and the inherent environment lighting and mobile user dynamics.
A new opportunity arises recently with the advancements in mobile 3D vision, including built-in high-accuracy depth sensors and deep learning-powered algorithms, which provide the means to better sense and understand the physical surroundings.
Centering the key idea of 3D vision, in this work, we design
an edge-assisted framework called \sysname to provide
mobile AR applications the ability to obtain accurate omnidirectional lighting estimation in real time.

Specifically, we develop a novel sampling technique that efficiently compresses the raw point cloud input generated at the mobile device. This technique is derived based on our empirical analysis of a recent 3D indoor dataset and plays a key role in our 3D vision-based lighting estimator pipeline design.
To achieve the real-time goal, we develop a tailored GPU pipeline for on-device point cloud processing and use an encoding technique that reduces network transmitted bytes.
Finally, we present an adaptive triggering strategy that allows \sysname to skip unnecessary lighting estimations and a practical way to provide temporal coherent rendering integration with the mobile AR ecosystem.
We evaluate both the lighting estimation accuracy and time of \sysname using a reference mobile application developed with \sysname's APIs. Our results show that \sysname takes as fast as 20.67ms per lighting estimation and achieves 9.4\% better estimation accuracy than a state-of-the-art neural network.
\end{abstract}

\maketitle

\section{Introduction}
\label{sec:intro}

\begin{figure*}[t]

\centering
    \begin{subfigure}[b]{0.485\linewidth}
        \centering
        \includegraphics[width=\linewidth]{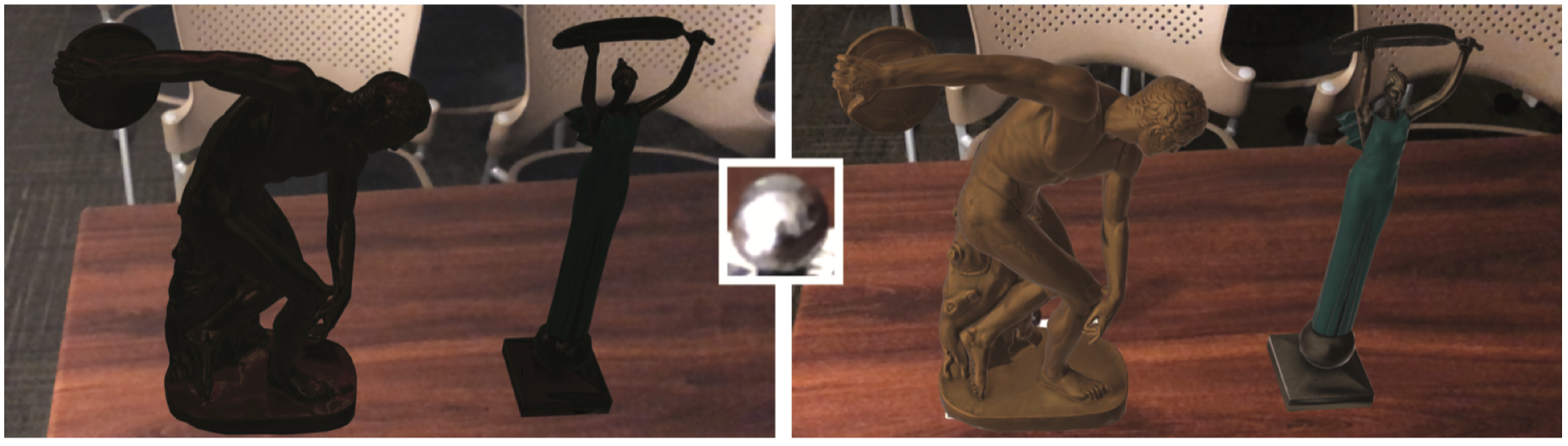}
        \caption{Left: ARKit vs. right: GLEAM (obtained directly from~\cite{prakash2019gleam})}
    \end{subfigure}
    \hfill
    \begin{subfigure}[b]{0.485\linewidth}
        \centering
        \includegraphics[width=\linewidth]{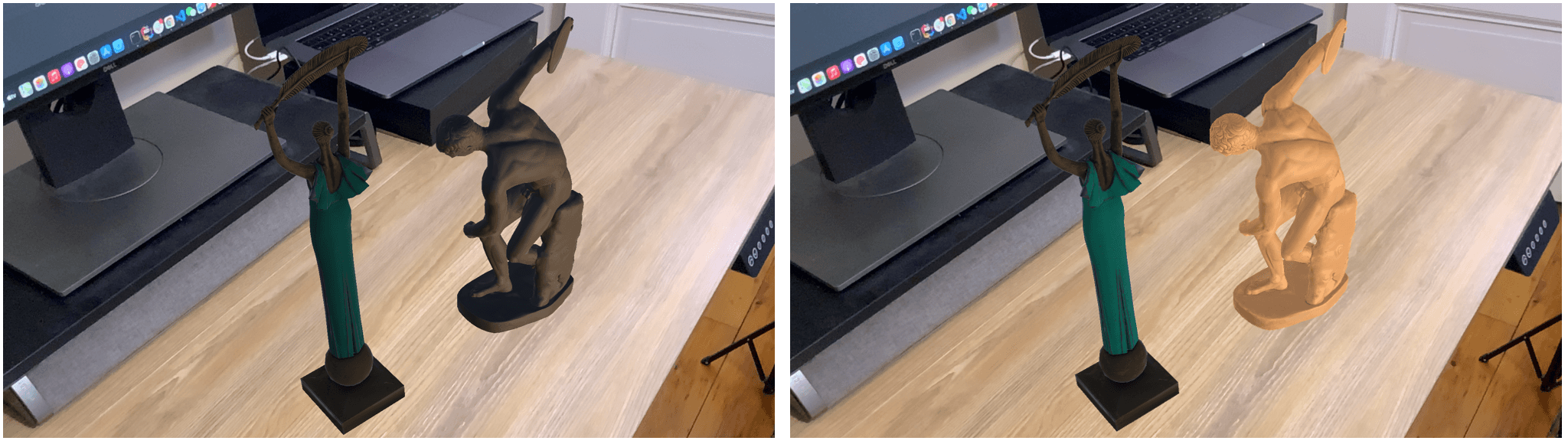}
        \caption{Left: ARKit vs. right: \sysname}
        \label{subfig:rendered_scenes:xihe}
    \end{subfigure}
\vspace{-0.5em}
    \caption{Rendered AR scenes.
\textnormal{Both GLEAM and \sysname achieve better visual effect compared to ARKit.
\sysname better captures the spatially-variant lighting difference without needing the physical probe, compared to GLEAM~\cite{prakash2019gleam}.
    Note we compared to ARKit's ambient light sensor based lighting estimation.
}
}
    \label{fig:rendered_scenes}
    \vspace{-1.5em}
\end{figure*}

Augmented reality (AR), overlaying virtual objects in the user's physical surrounding, has the promise to transform many aspects of our lives, including tourism, education, and online shopping~\cite{Inter_IKEA_Systems_B_V2017-iv,Google_for_Education_undated-rj}.
The key for AR to success in these application domains heavily relies on the ability of \emph{photorealistic rendering}, a feature which can be achieved with access to omnidirecitional lighting information at rendering positions~\cite{debevec2006image}.
For example, a virtual table should ideally be rendered differently depending on the user-specified rendering positions---referred to as \emph{spatially-variant rendering}, to more accurately reflect the environment lighting and more seamlessly blending the virtual and physical worlds.

However, obtaining such lighting information necessary for spatially-variant photorealistic rendering is challenging in mobile devices.
Specifically, even high-end mobile devices such as iPhone 12 lack access to  $360^\circ$ panorama of the \emph{rendering position}. Even though with explicit user cooperation, it is possible to obtain the $360^\circ$ panorama of the \emph{observation position} via the use of ambient light sensors and front-/rear-facing cameras.
Directly using the lighting information at the observation position, i.e., where the user is at, to approximate the lighting at the rendering position, i.e., where the virtual object will be placed, can lead to undesirable visual effects due to the inherent lighting spatial variation~\cite{Gardner2017}.

One promising way to provide accurate omnidirectional lighting information to mobile AR applications is via 3D vision support.
With the recent advancement in mobile 3D vision including built-in high-accuracy Lidar sensors~\cite{ipad-lidar} and low-complexity high-accuracy deep learning models~\cite{pointar_eccv2020,qi2016pointnet,Wu_2019_CVPR}, we are bestowed upon a new opportunity to provide spatially-variant photorealistic rendering!
In this work, we design the first 3D-vision based framework \sysname that provides mobile AR applications the ability to \emph{obtaining accurate omnidirectional lighting estimation in realtime}.
Our design can be broadly categorized into three parts: \1 algorithm and system design to support spatially-variant estimation; \2 per-frame performance optimization to achieve the real-time goal; and \3 multi-frame practical optimization to further reduce network cost and to integrate with existing rendering engines for temporal coherent rendering.
We implement the framework \sysname on top of Unity3D and AR Foundation as well as a proof-of-concept reference AR application that utilizes \sysname's APIs.
Figure~\ref{fig:rendered_scenes} compares the rendered AR scenes using \sysname and prior work~\cite{prakash2019gleam}.

To support the key goal of spatially-variant lighting estimation, we design an end-to-end pipeline for 3D data processing, understanding, and management.
Specifically, we devise a novel sampling technique called \emph{\uspc} technique to preprocess raw 3D data in the format of point cloud.
This technique is derived based on our empirical analysis using a recent 3D indoor dataset~\cite{pointar_eccv2020}; our analysis shows the correlation between the incomplete observation data (i.e., not $360^\circ$ panorama) and the lighting estimation accuracy.
Further, we redesign a recently proposed 3D vision-based lighting estimation pipeline~\cite{pointar_eccv2020} by leveraging our \uspcs technique to transform raw point clouds to compact representations while preserving the observation completeness.
To better support mobile devices of heterogeneous capacity and simplify the client design, we centralize the tasks, including point cloud and lighting inference management, into a stateful server design.
Our edge-assisted design also facilitates sharing among different mobile users and therefore provides opportunities to improve lighting estimation with \emph{extrapolated point cloud data}, e.g., via merging and stitching different observation data to increase the completeness.

To achieve the real-time goal, we develop a tailored GPU pipeline for processing point clouds on the mobile device and use an encoding technique that reduces network transmitted bytes.
Specifically, we leverage the property of point cloud in which the computation for each point can be parallelized and devise a strategy that trade-offs storage to improve runtime performance. In essence, we pre-generate densely sampled sphere coordinates in the \uspc and pre-compute their distances to \emph{anchors}---a set of uniformly distributed surface points in a unit sphere.
At runtime, instead of trying to search for the closest anchor to each projected point, we simply search within the densely sampled sphere coordinates.
We refer to these densely sampled sphere coordinates as \emph{acceleration grids}.
Further, we encode each colored anchor of the \uspc with unsigned 8bits int and half-precision 16bits float where appropriate based on the practical characteristics such as common image format of LDR and the depth sensor precision.
Our \uspc design also allows easy stripping of unnecessary data by removing any uninitialized anchors ,i.e., anchors that are not colored due to incomplete raw observation point cloud.

Finally, we present an adaptive triggering strategy that allows \sysname to skip unnecessary lighting estimations and a practical way to provide temporal coherent rendering integration with the AR ecosystem.
The key idea of the triggering strategy is to leverage an easy-to-obtain and fast-to-compute metric to determine directly on the mobile device whether the lighting condition has changed sufficiently to warrant a new lighting estimation at the edge. We use a sliding window-based approach that compares the \uspc changes between consecutive frames.
To achieve temporal-coherent visual effects, we leverage additional mobile sensors including ambient lighting and gyroscope to better match the lighting estimation responses with the current physical surroundings.
We also detail steps to leverage a popular rendering engine to apply the spatially-variant lighting on virtual objects.

Spatially-variant lighting information can be traditionally extracted using physical probes~\cite{debevec2006image,prakash2019gleam}, and more recently estimated with deep neural networks~\cite{Song2019,Garon2019,Gardner2017,pointar_eccv2020}.
For example, Debevec et al. demonstrated that \svl can be effectively estimated by using reflective sphere light probes to extrapolate camera views.
More recently, Prakash et al. developed a mobile framework that provides real-time lighting estimation using physical probes~\cite{prakash2019gleam}.
On a different vein, new deep learning-based approaches that do not require the use of physical probes have demonstrated efficiency in estimating \svl. The early efforts mostly focus on model innovation but still incur high computational complexity, making them ill-suited to run on mobile devices~\cite{Song2019,Garon2019,Gardner2017}. Until very recently, Zhao et al. proposed a lightweight 3D vision-based approach that takes advantage of new mobile depth sensors and shows promise of being mobile-friendly~\cite{pointar_eccv2020}.
Our work leverages the advancement of mobile 3D vision and presents the first framework that supports accurate omnidirectional lighting estimation in real time via algorithm and system co-design. Moreover, our work does not require the use of physical reflective probes at runtime, thus can support a wider range and more practical AR application scenarios.

Our main contributions of this paper are:

\begin{itemize}[leftmargin=.12in,topsep=4pt]
    \item We design and implement a 3D vision-based framework \sysname that allows mobile AR applications to obtain spatially-variant lighting estimation and to achieve temporal-coherent rendering, fast and accurately. The relevant research artifact is available at: \textcolor{magenta}{\url{https://github.com/cake-lab/Xihe}}.
    \item We propose a novel point cloud sampling technique that effectively compresses the observation point cloud without impacting the estimation accuracy. This sampling technique is used in conjunction with a lightweight neural network to provide the spatially-variant lighting estimation.
    \item To achieve the real-time goal, we propose two per-frame optimizations, namely a tailored GPU pipeline for point cloud operations on the mobile devices and a practical data encoding scheme. This allows \sysname generate lighting estimations as fast as 20.67ms. We design an adaptive triggering strategy that effectively reduces up to 76.24\% estimation requests by allowing \sysname client to identify lighting condition changes directly on the mobile devices.
    \item We conduct a comprehensive evaluation with a real-world 3D indoor RGB-D dataset on several mobile devices and network conditions and show that \sysname can achieve better visual effects than two existing approaches, i.e., GLEAM~\cite{prakash2019gleam} and ARKit~\cite{arkit}.
    Further, we also present a detailed performance breakdown of \sysname under different configurations and use cases, reporting mobile, network, and edge time. Our study reveals a number of important factors such as number of anchors and estimation positions. Lastly, our lab-based evaluation showcases \sysname's ability to effectively generate lighting estimations by adapting to both environment lighting and user movement dynamics.
\end{itemize}

\section{Problem and Solution Overview}

In this paper, we set out to address the key problem of providing spatially-variant photorealistic rendering in the context of mobile AR.
\emph{Photorealistic rendering} at a high level is about displaying lifelike virtual 3D objects which mobile users cannot easily distinguish from physical objects.
The key challenge to achieving photorealistic rendering lies in obtaining the omnidirectional lighting of a geometric space where the virtual objects will be displayed.
Larger objects therefore require more lighting information at different points in the geometric space.
The geometric center of the virtual object, referred to as the \emph{placement position}, is often assumed to be specified by the user at runtime.
A placement position can be extrapolated to multiple \emph{estimation positions} where each position corresponds to a lighting representation, e.g., \SHc.
The advent of \emph{3D vision}, the ability to perceive both color and depth information, creates a new opportunity to transfer the measurable lighting information at the observation position to the placement position i.e. rendering positions.

\para{Challenges.}
We leverage the key idea of 3D vision, and address three key challenges in designing a 3D vision-based lighting estimation framework called \sysname.
The first challenge lies in accurately representing the spatially-variant lighting given the inherent constraints of mobile AR scenarios including limited field-of-view, user mobility, and environment lighting changes.
The second challenge is to provide such accurate lighting estimation fast enough so that rendering engines can utilize this information for each frame if necessary.
The third challenge is to provide temporal-coherent rendering that considers cross-frame visual harmony when utilizing estimated lighting information.

\para{Solution Overview.}
In this work, we design a 3D-vision based framework \sysname that provides mobile AR applications the ability to \emph{obtaining accurate omnidirectional lighting estimation in real time}.
See Figure~\ref{fig:architecture} for an architecture design of \sysname.
Our design can be broadly categorized into three intertwined parts. We first introduce the algorithm and system design to support spatially-variant estimation in Section~\ref{sec:supporting_svle} and then describe our per-frame performance optimization to achieve real-time goal in Section~\ref{subsec:per_frame_optimization}.
We further describe our cross-frame practical optimization to reduce network cost and to integrate with a popular rendering engine for temporal-coherent rendering in Section~\ref{subsec:multi_frame_optimization}.

\section{Spatially-variant Estimation}
\label{sec:supporting_svle}

Spatially-variant lighting allows representation of lighting at different world positions.
As such it  has the promise to provide more photorealistic renderings of virtual objects, which is especially important in the realm of augmented reality.
For example, when used in a furniture shopping app, spatially-variant lighting can more effectively render a piece of couch with different outlooks depending on the user's physical environment (well-lit room or darker room), the rendering position, and the couch size.
Figure~\ref{fig:rendered_scenes} visualizes rendering examples with and without spatially-variant lighting information.

\begin{figure}[t]
    \centering
    \begin{subfigure}[t]{0.4\linewidth}
        \includegraphics[width=\linewidth]{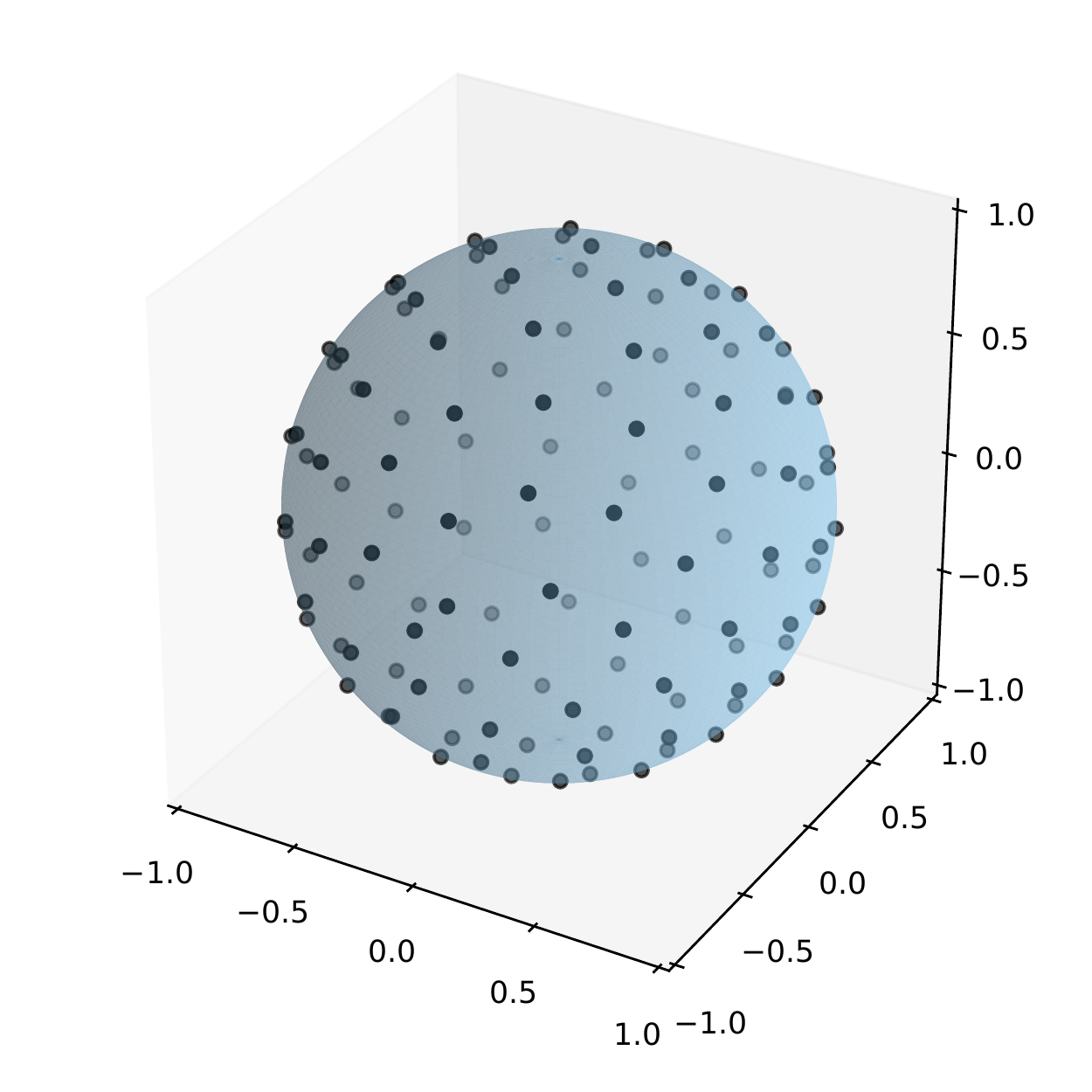}
        \caption{Uniform distribution.}
        \label{fig:uniform_points}
    \end{subfigure}
    \quad
    \begin{subfigure}[t]{0.4\linewidth}
        \includegraphics[width=\linewidth]{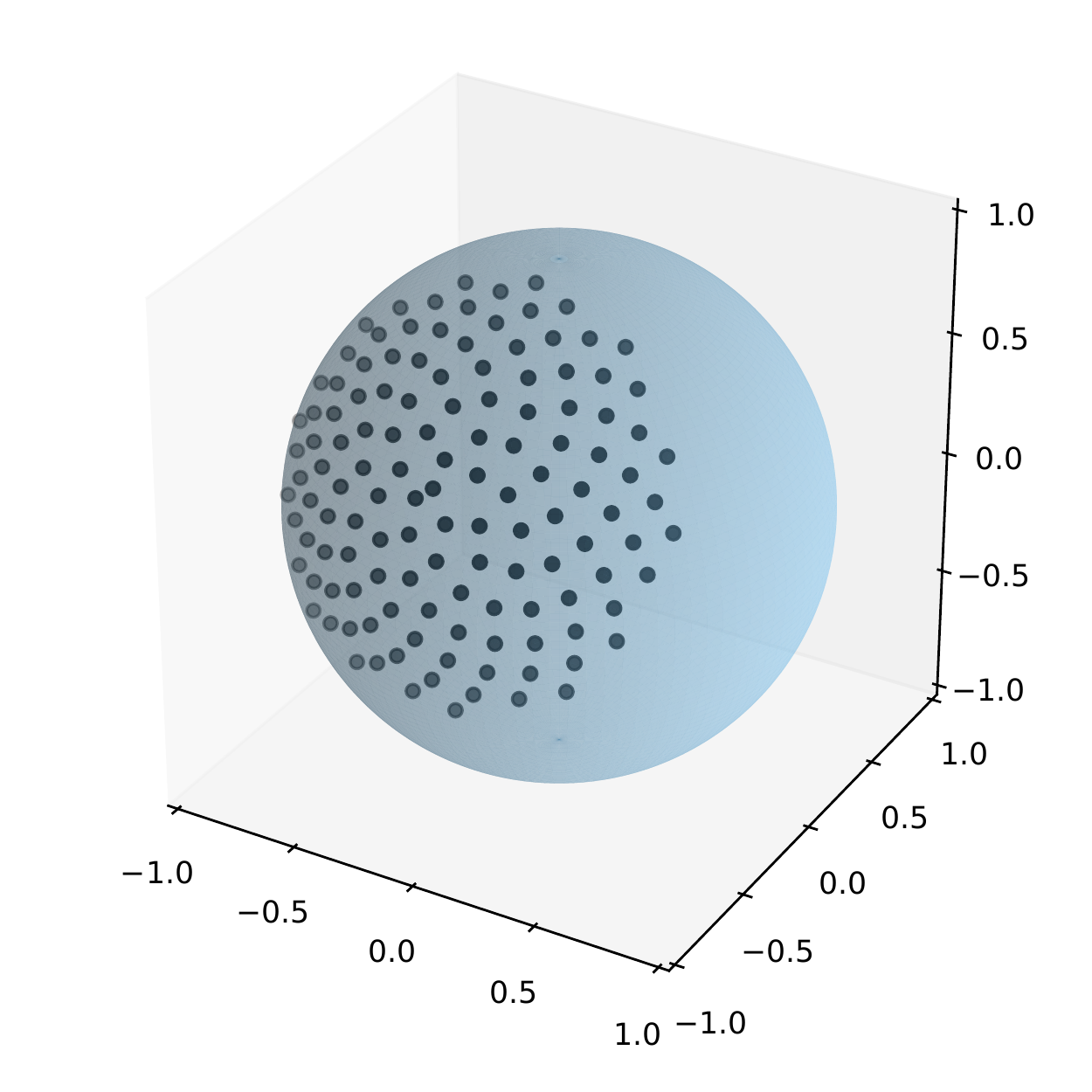}
        \caption{Non-uniform distribution.}
        \label{fig:grouped_points}
    \end{subfigure}
    \vspace{-0.5em}
    \caption{Impact of anchor point distributions on observation completeness.
    }
    \label{fig:sphere_distribution}
\end{figure}

\subsection{Unit-Sphere based Point Cloud Sampling}
\label{sec:uspc}

\sysname enables real-time efficient lighting estimation for mobile devices with 3D vision-based deep learning model that takes point cloud generated from on-device camera captured RGB-D image as input.
Such point cloud data is usually large in size and can contain redundant information~\cite{Schnabel2006-kg}. Therefore, down-sampling point cloud is beneficial to computation and network efficiency.
However, directly down-sampling the raw point cloud using techniques such as uniform sampling can negatively impact lighting estimation accuracy.
In this section, we present our novel \uspcs technique which preserves observation field-of-view (FoV) as much as possible. Our design is informed by an empirical analysis that demonstrates the negative correlation between observation completeness and lighting estimation accuracy.

\subsubsection{Impact of Incomplete Observation Data}
\label{subsubsec:completness_impact}

To study the potential impact of incomplete observation data on lighting estimation accuracy, we first propose an entropy-based metric to measure the point cloud observation completeness.
We define the \emph{observation completeness} as the percentage of colored anchors when projecting a point cloud to unit sphere surface.
Further, \emph{anchor points} are defined as a set of uniformly distributed surface points in a unit sphere $O$.
Intuitively, the observation completeness depends on both the points distribution ($D$) and anchor distribution ($A$).
We define the joint entropy $H(D, A)$ as:

\begin{equation}
    {\displaystyle \mathrm {H} (D, A)=-\sum_{i \in S}\sum_{j=1}^{i}{\mathrm {P} (d_{ij}, a_i)\log_2 [\mathrm {P} (d_{ij}, a_i)}}].
    \label{eq:entropy}
\end{equation}

where $P(d_{ij}, a_i)$ is the joint probability of projecting points into a unit sphere with $i$ anchor point, and $S$ is a set of possible anchor sizes (i.e., number of anchor points). In this work, we choose $S = \{2^k | 1 \leq k \leq 12\}$.
By using the Equation~\eqref{eq:entropy}, we can succinctly measure the point cloud observation completeness. The higher the entropy value, the more complete the observation. Additionally, it allows us to easily distinguish the observation completeness for point clouds of the same size.
For example, even though the two projected point clouds have the same number of points, the projected point cloud shown in Figure~\ref{fig:uniform_points} is considered to be more complete than the one shown in Figure~\ref{fig:grouped_points}.

Next, we leverage a recently proposed state-of-the-art 3D vision based lighting estimation algorithm and its point cloud dataset~\cite{pointar_eccv2020} to correlate observation completeness with estimation accuracy.
The raw point clouds, each has around 82K points, were first uniformly downsampled to 1280 points.
We then trained a model (see Section~\ref{subsec:eval_3d_estimator} for training setup details) based on the original paper's description.
Lastly, we obtained the lighting estimation error, represented as Mean Squared Error (MSE), by evaluating the trained model on each point cloud and compare the results to ground truth.

\begin{figure}[t]
    \centering
    \begin{subfigure}[b]{0.48\linewidth}
        \centering
        \includegraphics[width=\linewidth]{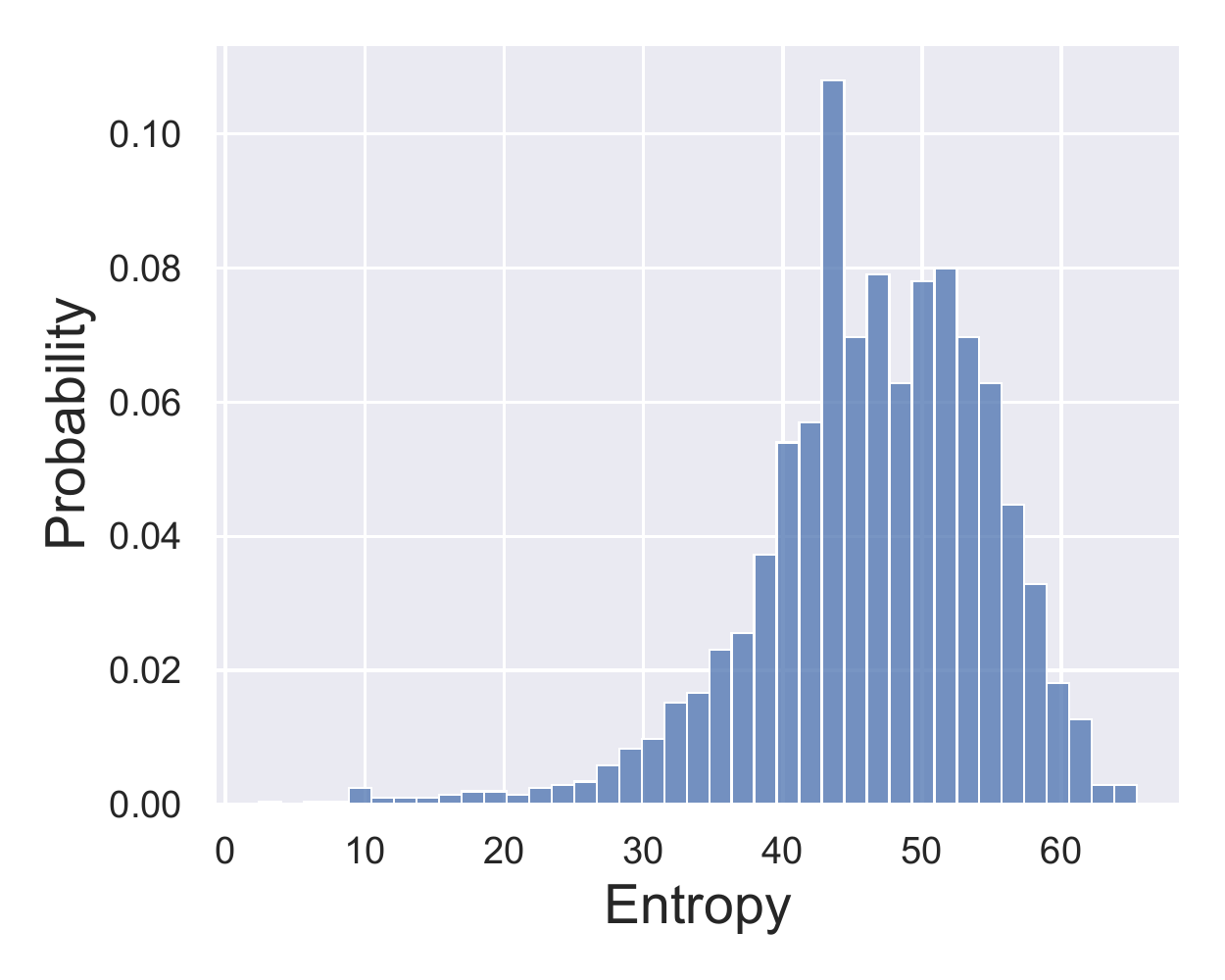}
        \caption{Entropy distribution.}
        \label{subfig:entropy_distribution}
    \end{subfigure}\hfill
    \begin{subfigure}[b]{0.48\linewidth}
        \centering
        \includegraphics[width=\linewidth]{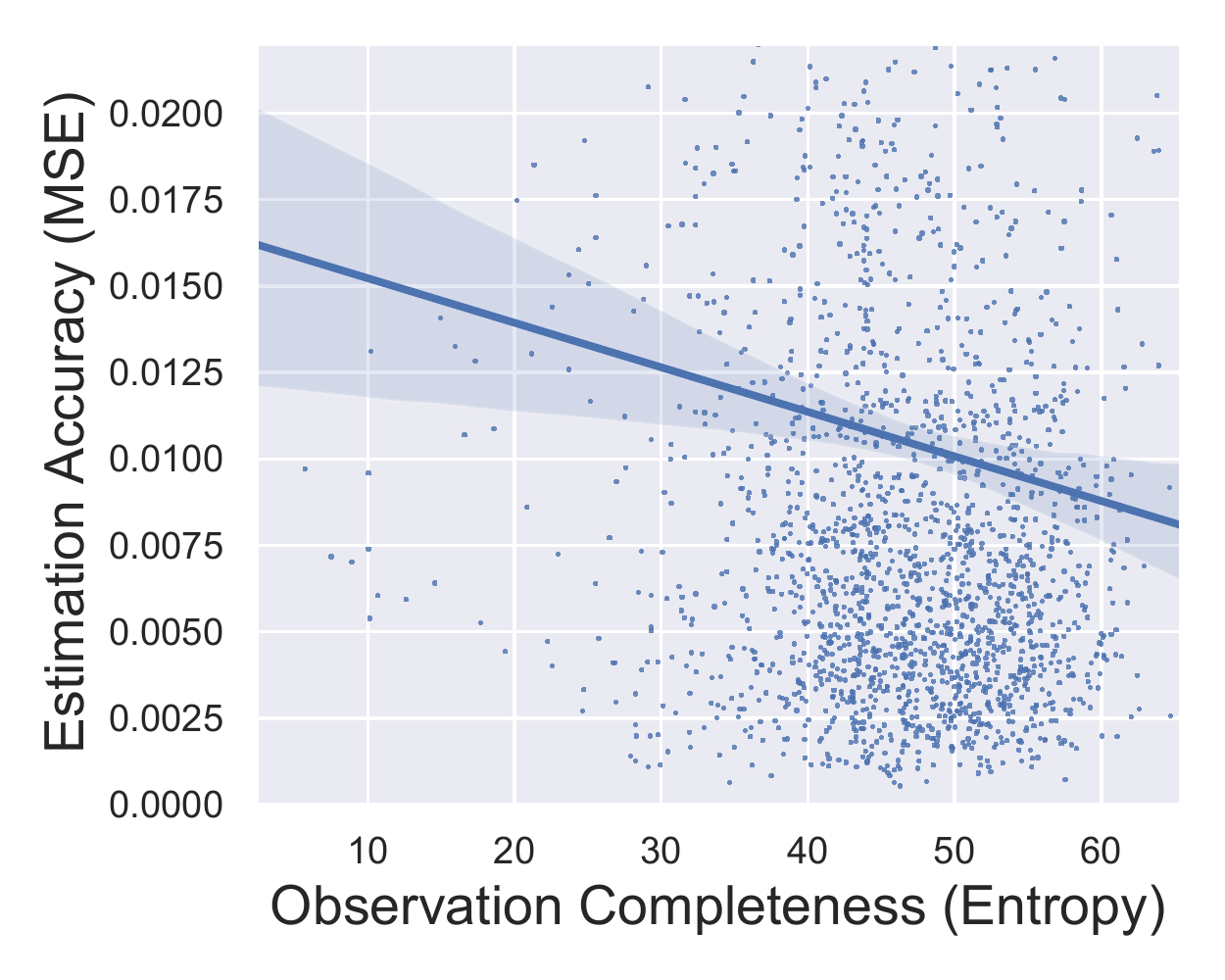}
        \caption{Entropy vs. accuracy.}
        \label{subfig:accuracy_entropy}
    \end{subfigure}
    \vspace{-1em}
    \caption{Observation-preserving metric empirical analysis.}
    \label{fig:entropy_analysis}
    \vspace{-1.5em}
\end{figure}

Figure~\ref{subfig:entropy_distribution} depicts the entropy distribution of all raw point clouds.
Figure~\ref{subfig:accuracy_entropy} shows a negative correlation between the observation completeness measured by $e$ and estimation accuracy.
In conclusion, we empirically observed that the coverage observation completeness is the key impact factor to lighting estimation accuracy.

\subsubsection{Observation Point Cloud Downsampling}
\label{subsec:how_to_downsample}

\begin{figure*}[t]
    \centering
    \begin{subfigure}[b]{0.19\linewidth}
        \centering
        \includegraphics[width=\linewidth]{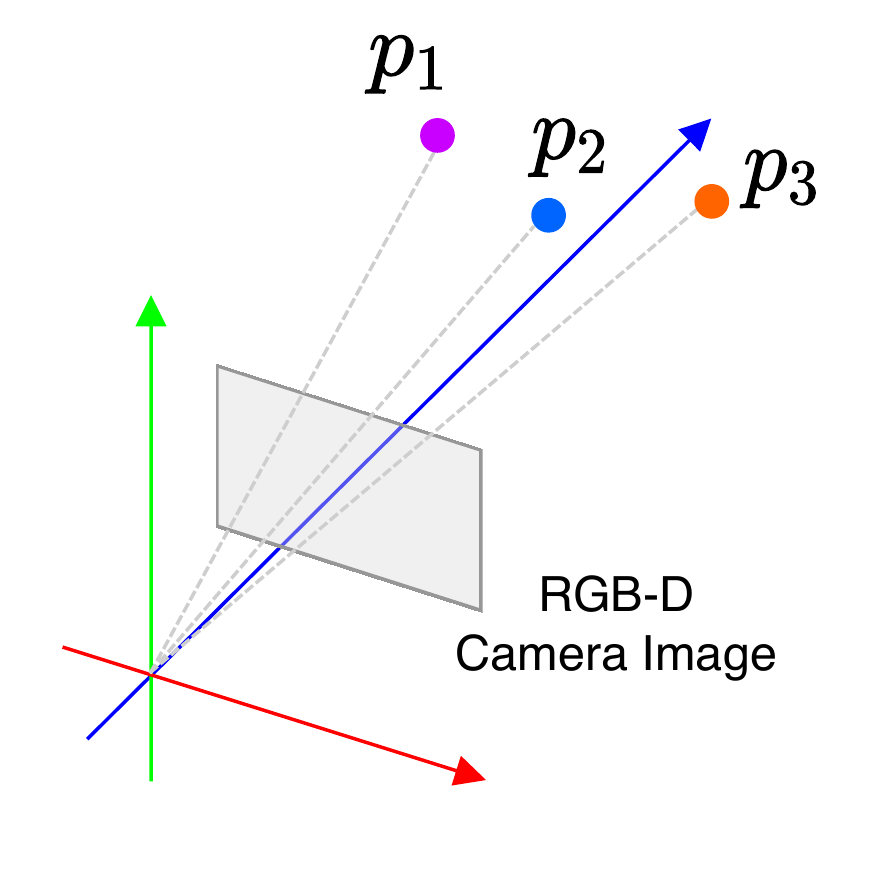}
        \caption{Camera input.
}
    \end{subfigure}
    \begin{subfigure}[b]{0.19\linewidth}
        \centering
        \includegraphics[width=\linewidth]{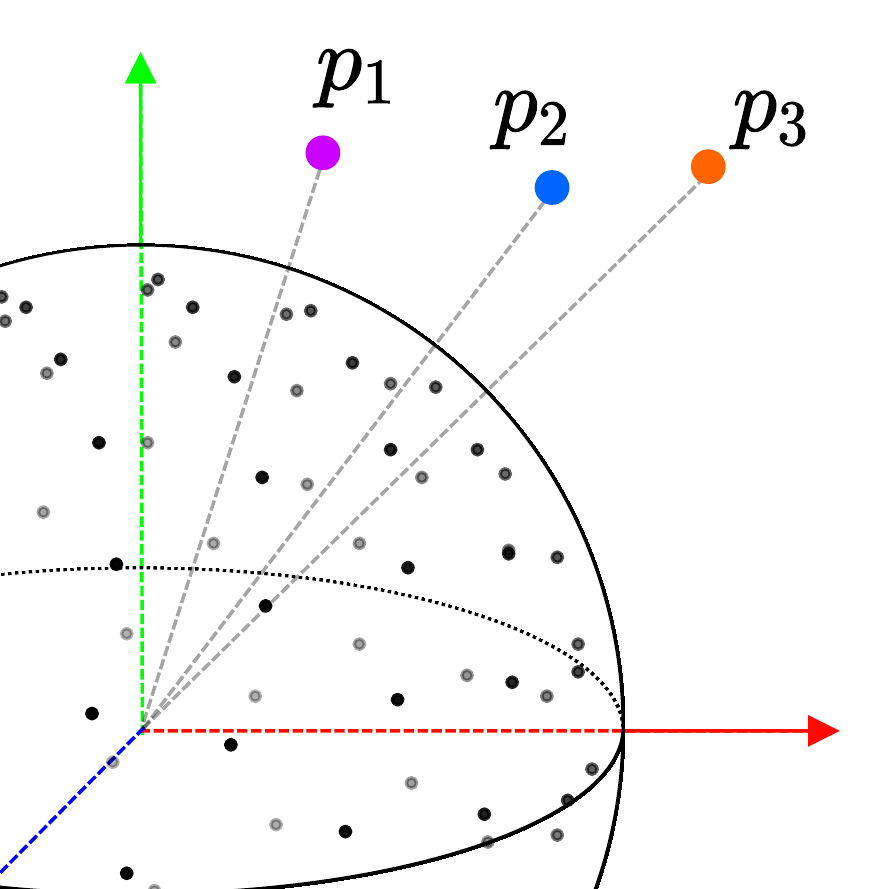}
        \caption{Raw points.}
    \end{subfigure}
    \hfill
    \begin{subfigure}[b]{0.19\linewidth}
        \centering
        \includegraphics[width=\linewidth]{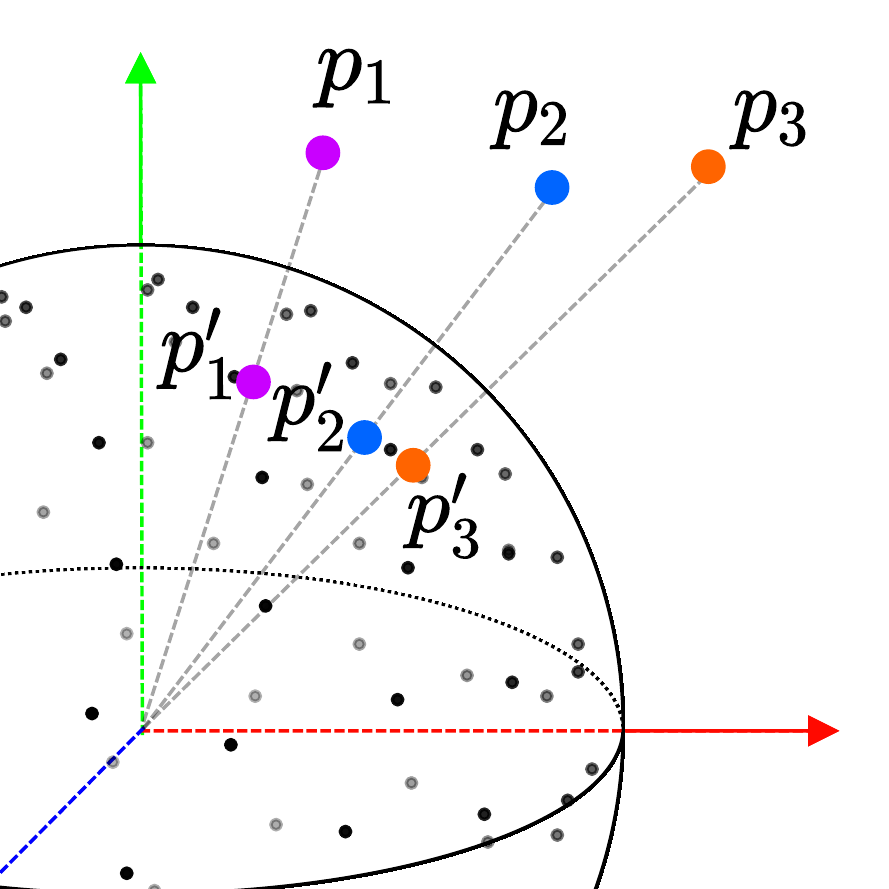}
        \caption{Point projection.}
    \end{subfigure}
    \hfill
    \begin{subfigure}[b]{0.19\linewidth}
        \centering
        \includegraphics[width=\linewidth]{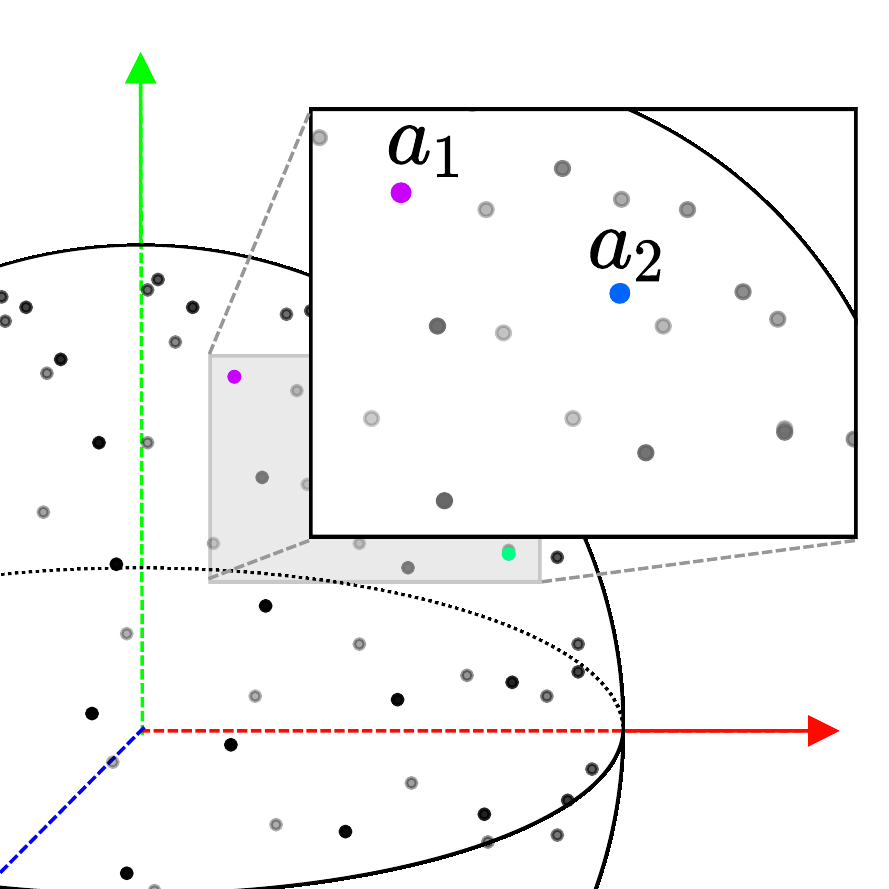}
        \caption{Anchor initialization.}
    \end{subfigure}
    \hfill
    \begin{subfigure}[b]{0.19\linewidth}
        \centering
        \includegraphics[width=\linewidth]{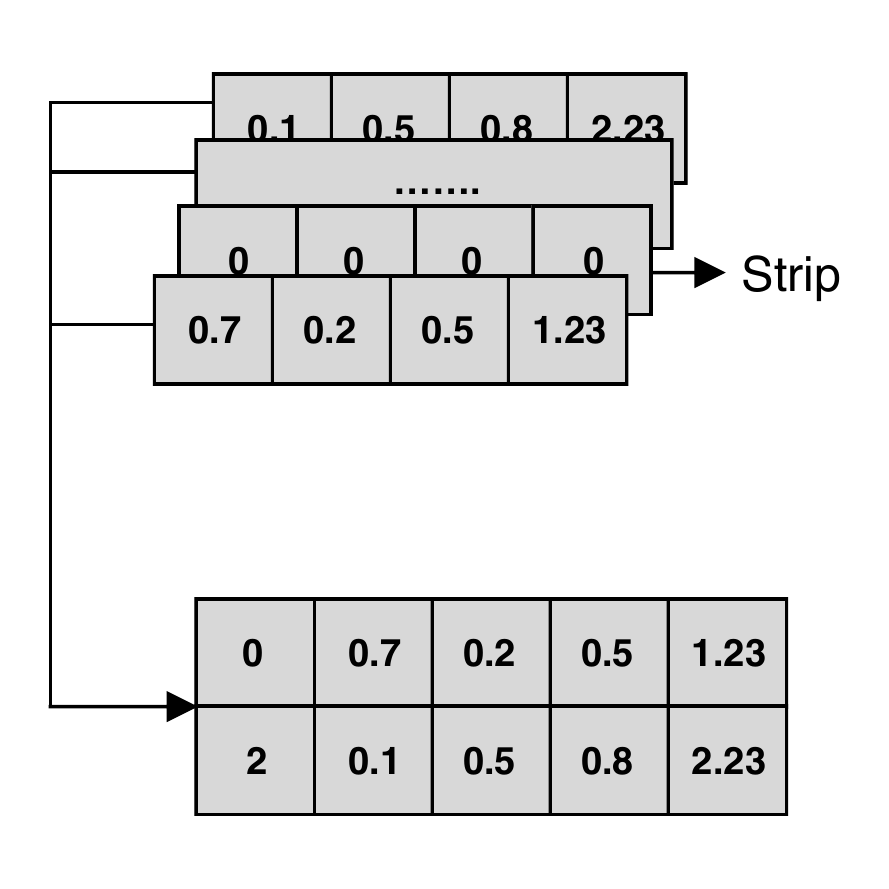}
        \caption{Encoding.}
    \end{subfigure}
    \vspace{-1em}
    \caption{An example of \uspc sampling and encoding.
}
    \label{fig:example_pcs}
    \vspace{-1em}
\end{figure*}

To effectively sample point cloud, we design a novel point cloud sampling method called \uspcs.
It first projects every point in a point cloud $P$ onto a unit sphere $O$, defined as $f(P,O)$, then find the nearest anchor point for each projected point.
By establishing the relationship between each point from $P$ and $O$, $f$ outputs a point cloud distribution $D$.
Finally, we downsample $P$ with a nearest point selection that approximates the depth culling process~\cite{depthbuffer}. That is, $P$ is downsampled by selecting the nearest points, i.e., points with the shortest 3D Euclidean distance to the sphere origin, from $D$ to color corresponding anchor points in $O$.

Figure~\ref{fig:example_pcs} provides an example walkthrough of our \uspc sampling. Considering three points ($p_1$, $p_2$, $p_3$) in the original point cloud, and their corresponding projected points ($p'_1$, $p'_2$, $p'_3$).
Each projected point is then matched to an anchor that is closest to itself.
We refer to these matched anchors as \emph{nearest neighbors}.
In this example, $p'_1$ will be matched to $a_1$ while $p'_2$ and $p'_3$ will both be matched to $a_2$.
Without loss of generality, if $p'_1$ is the \emph{only} projected point that $a_1$ is assigned as the nearest neighbor, then $a_1$ will be initialized with the RGB values of $p'_1$ and $D(p'_1, o)$---the distance between $p'_1$ and the sphere origin $o$.
Similarly, if $p'_2$ and $p'_3$ are the only projected points that $a_2$ is assigned, and $p'_2$ is closer to the sphere origin than $p'_3$ is, $a_1$ will be initialized with the RGB values of $p'_2$ and $D(p'_2, o)$.

In practice, the points set $P$ is determined at runtime by configurations and camera hardware while the size of $O$ is a configurable system parameter; the ratio $\frac{size(P)}{size(O)}$ represents the down-sampling ratio.
In other words, we will leverage the \uspc as the basis for estimating the spatially-variant lighting information instead of directly using the observed point cloud.
In an AR session, we need to perform several consecutive \uspcs within a small time span for each estimation position, as will be described in section~\ref{subsec:triggering}.
The \uspc for each estimation position will likely only be partially initialized, due to incomplete environment observation.

We will store the \uspc in a custom designed data structure, represented as a 4D vector (the RGB values and the 3D Euclidean distance between the projected points and the sphere origin).
This data structure design has two major advantages.
First, sphere anchor positions can be pre-computed ahead of time.
As such, our data structure only needs to store anchors in an ordered array with each index corresponds to an anchor position.
This design also presents an opportunity to speedup \sysname's triggering strategy using pre-computed neighbors for each anchor, as will be described in Section~\ref{subsec:triggering}.
Second, storing \uspc using this data structure allows \sysname to extract both 3D space positions and colors at viewing directions for estimation positions.

\subsubsection{Downsampled Point Clouds for Virtual Objects}
\label{subsec:how_to_downsample_for_objects}

\sysname supports estimation positions that are specified via \sysname's APIs and can perform \uspcs at each estimation position.
Additionally, \sysname also provides a simplified workflow that automatically assigns estimation positions when a virtual object is placed to the scene.
Specifically, given a virtual object's placement position (e.g., specified by the user), \sysname will first designate the placement position as one estimation position and subsequently generate multiple estimation positions based on the object size.
In other words, \sysname can support multiple lighting estimation requests, the response of each is represented as \SHc, for a given object.
However, due to current mobile rendering engine limitations, e.g., in the case of mobile Unity3D, \sysname will only render each object with one set of \SHc.
To circumvent this limitation, AR developers using \sysname can either decompose the large object into smaller ones or customize rendering shaders to take advantage of multiple sets of \SHc.

\subsection{3D Vision-based Estimation Pipeline}
\label{subsec:3d_vision_based_estimator}

Another key design to support spatially-variant lighting comes down to an efficient algorithm that can extract and estimate lighting information from an incomplete environment point cloud.
As depth sensors such as Lidar start to be equipped with mobile devices, it is now possible to leverage 3D vision-based algorithms for Mobile AR applications.

However, given that mobile 3D vision is in its infancy, there have been very few works in mobile-friendly 3D vision-based lighting estimation techniques~\cite{pointar_eccv2020,rohmer2017natural}.
We choose the state-of-the-art 3D vision based lighting estimation model PointAR~\cite{pointar_eccv2020} as the building block for designing a new lighting estimation neural network that works well with \uspc.
Briefly, PointAR is a two-staged neural network pipeline that consists of:
\1 a point cloud transformation to simulate the camera movement from the observation position to the rendering position;
\2 a point cloud-based compact deep learning model.
Our practical \xiheNet is designed by integrating our novel \uspcs technique, as described in Section~\ref{subsec:how_to_downsample}, into the first stage.

To train the \xiheNet model, we first generate six training/test datasets---with the following number of anchors: [512, 768, 1024, 1280, 1536, 2048]---each consists of 608k/2037 \uspc from the PointAR RGB-D dataset.
Then, we extract the ground truth lighting information, represented as \SHc from both LDR and HDR environment maps.
The resulting data item is in the form of a downsampled point cloud and the corresponding \SHc.
For LDR-based and HDR-based datasets, we train six instances of \xiheNet to study the performance and estimation accuracy trade-offs.
As we later observe that \xiheNet models trained with LDR-based datasets lead to better visual effects than that of HDR-based datasets;
for the remainder of the paper, we will report results using \xiheNet trained on LDR-based datasets.

\sysname outputs \SHc as an omnidirectional representation of environment lighting at a single world position for rendering.
If directly using image-based lighting estimation models~\cite{Garon2019,Song2019}, one needs to post-process to correctly orient estimated \SHc since the 3D world orientation cannot be represented on the image input.
Our \xiheNet guarantees the orientation constant~\cite{pointar_eccv2020} by explicitly considers the world space point cloud and estimates \SHc at the same orientation.

\subsection{Edge-assisted Resource Sharing}
\label{subsec:edge_management}

\subsubsection{Point Cloud Management}

Naively preserving and managing time-series downsampled point clouds on mobile devices can be harmful to overall system performance as it can consume too much device memory.
Therefore, \sysname proposes to use a stateful edge-based point cloud management design.
Moreover, we abstract \sysname client as a point cloud provider to handle mobile heterogeneity, like camera parameters, through adapters.

We design the \sysname server to manage the \uspc for each estimation position.
Edge data will be updated throughout the AR session as new lighting estimation requests come in.
Currently our \xiheNet can produce highly accurate estimations even with per-frame data that have a number of uninitialized anchors.
Based on our empirical analysis between observation completeness and the estimation accuracy, one practical way to further improve the estimation accuracy is to leverage multiple frames to progressively complete the \uspc, i.e., with more colored anchors.
Thus, \sysname server manages the collected environment point cloud in an accumulative fashion and merges the point cloud with data associated with newly triggered requests.
As such, we can then augment the \uspc sent with each estimation request with historical data for improving observation completeness.
Such augmentation is particularly useful when environment observation data is shared among multiple estimation positions or clients in the same AR session.

\subsubsection{Lighting Estimation Management}
Lighting estimation is inherently latency sensitive due to the strict latency requirement.
In the context of mobile AR, rendering engine typically targets 30fps refresh rate, which corresponds to refresh approximately 33.33ms per frame.
Ideally lighting estimation should be performed at each frame to achieve best accuracy.
However, for deep learning-based estimation algorithms, fulfilling such requirement on mobile devices can be very challenging given the resource constraints and the likelihood of sharing on-device computation resources with tasks like rendering.

To address the low latency requirement and better support our overall vision of supporting multi-users AR sessions and multi-objects rendering, we design \sysname to run the inference execution on a GPU-accelerated edge server.
A strategically provisioned edge can provide a low-latency and high-bandwidth connection between the mobile devices and the server~\cite{modi}, as well as the potential to batch inference requests.
By using edge-based deep learning inference, \sysname has the potential to support more complex model and achieve much lower computation latency, e.g., by leveraging the powerful edge resources.
The \sysname server will use the merged point cloud as the input to the 3D vision-based estimator and send back the estimation result in \SHc to \sysname client.

 \section{Fast and Accurate Estimation}
\label{sec:fast_and_accurate}

\begin{figure}[t]
  \includegraphics[width=\linewidth]{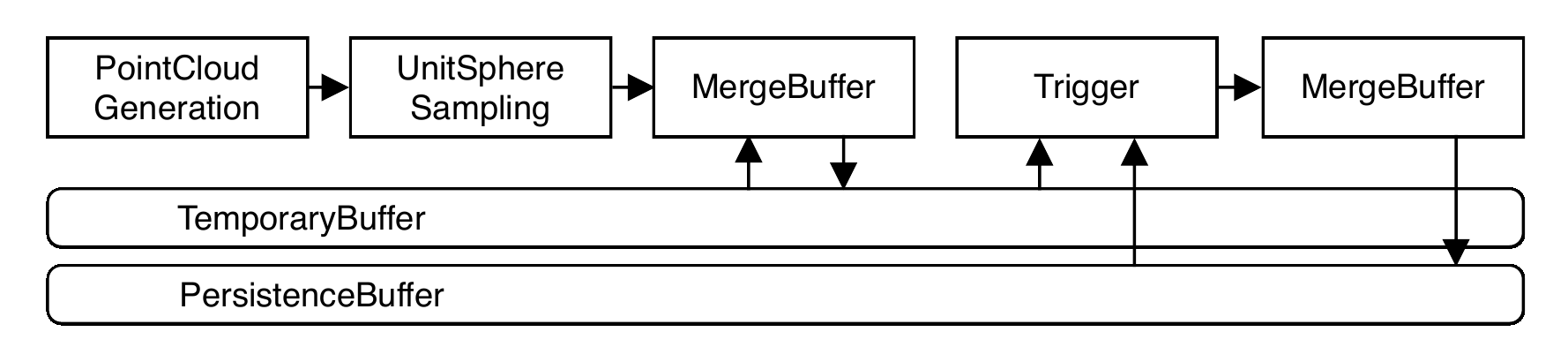}
  \vspace{-1.5em}
  \caption{\sysname mobile GPU processing pipeline.
}

  \label{fig:gpu_pipeline}
  \vspace{-1.5em}
\end{figure}

Continuously processing point cloud data directly on mobile CPU can be time consuming, thus might violating the real-time goal.
For example, a RGB-D image captured on a iPad Pro with the max resolution of 256x192 corresponds to 49k raw points.
Fortunately, operations on point clouds can naturally be parallelized at per-point level.
We explicitly assemble as many operations as possible to run on mobile GPU to reduce the CPU-GPU communication overhead for achieving the best performance; the processing pipeline is shown in Figure~\ref{fig:gpu_pipeline}.
Further, as constantly sending all the RGB-D data over the network to the edge can be costly and might not be necessary, we devise an adaptive triggering strategy that skips inference requests if the environment lighting conditions do not change substantially.
The triggering component (see Section~\ref{subsec:triggering}) decides whether \sysname should offload the point cloud sample to the edge sever for inference or store it into a temporary buffer.

\subsection{Per-frame Real-time Optimization}
\label{subsec:per_frame_optimization}

\subsubsection{Accelerating Point Cloud Sampling}
\label{subsec:tailored_gpu_pipeline}

Our processing pipeline, as shown in Figure~\ref{fig:gpu_pipeline}, starts with generating point cloud data from a camera captured RGB-D image feed, and performs generation at a pre-configured refresh rate.
By leveraging GPU computing, our point cloud generation code can be fully parallelized for common RGB-D image resolutions, e.g., 256x192 in the case of Lidar-equipped iPad.
Each generation outputs a point cloud of the world environment that is captured in the current camera view.

Then, we apply \uspcs to the generated point cloud.
However, naively using this sampling technique is computationally intensive, as it runs in $\Theta({\rm size}(P) * {\rm size}(O))$ time.
Although it is possible to parallelize both point processing and anchor searching, the number of required GPU threads can exceed the maximum support.
For example, if we were to fully parallelize the sampling process, i.e., using one GPU thread for searching each point-anchor pair, on a point cloud with 49k points and 1280 anchors, we will need about 62M GPU threads, while the maximum supported GPU thread group size on mobile Unity3D platform is 65535.
Although partially parallelizing the sampling, e.g., running the neareast anchor search for each point on a GPU thread, may comply with the current mobile GPU requirement, the execution time of each thread will be elongated.
Therefore, we propose an optimization method for reducing the computation resource requirement of this sampling technique.

Specifically, we propose to build a 2D \emph{acceleration grid} that uses densely pre-sampled points and pre-computed results to reduce the neareast anchor searching time.
Specifically, we first densely sample a set of points on the unit sphere based on quantizing the spherical coordinates polar angels $\theta$ and $\phi$.
Then, we pre-compute the nearest anchor for all the densely sampled points and store the corresponding anchor index in the acceleration grid.
At runtime, for each projected point, we first convert its cartesian coordinate to spherical coordinate and then apply the same quantization to match the point to a densely sampled point in the grid. The key advantage of doing so is that pre-sampled points can be stored as a 2D array, and indexed with spherical coordinates at runtime cheaply.

Note that using our proposed acceleration grid may introduce sampling errors as in essence this approach presents the entire sphere surface with discrete sampled points.
Intuitively, the more points the grid has, the better the approximation.
We empirically show that given a \uspc with 1280 anchors, using a pre-computed grid of 1024x512 points allow 97\% projected points match to their neareast anchors and only incur a negligible estimation error. More details will be presented in Section~\ref{subsec:impact_pcs}.
After a new \uspc data is sampled, \sysname client will continue the GPU pipeline execution by merging the newly generated data with historical data in the temporary buffer using an extrapolation operation.
The merging operation is an anchor-wise copy operation between two buffers, and will always overwrite old data with new one.

An alternative approach to accelerate \uspcs is to build a bounding volume hierarchy---a technique to accelerate ray tracing in real-time rendering~\cite{gunther2007realtime}---by properly subdivising the sphere surface to reduce the search space. However, leveraging such subdivision methods is nontrivial as one has to balance a number of factors such as the sphere surface dividing time, division access time, and total search time.
We leave this exploration as part of future work.

\subsubsection{Unit Sphere-based Point Cloud Encoding}
\label{subsec:network}

\sysname promises the low-latency network communication by leveraging above-mentioned compact point cloud data structure with byte-optimized encoding method.
When an estimation request is triggered, \sysname client sends the corresponding encoded \uspc as a byte-encoded HTTP packet to the \sysname server.
Our encoding consists of two steps: the striping step which removes all uninitialized anchors from the \uspc and the byte representation step which stores each point with fewer bits.

Specifically, instead of storing each point of the downsampled point cloud with four 32bits single precision floats, we use 8bits unsigned int and 16bits half-precision float to represent each point.
Though the original format is more precise to calculate and performs better as it aligns to the GPU bus transaction size, such data format uses redundant data bytes.
For example, when dealing with low dynamic range (LDR) camera images, 8bits data is usually sufficient to preserve the useful information.
Also, due to limitations such as depth sensor precision and camera visible area size, the distance information can be presented with 16bits half precision float.
Lastly, for each colored point, we use an extra 16bits to store their indices.

Our encoding scheme can lead to significantly savings both in terms of per-request and per-pixel data size.
For example, for a \uspc generated from a LDR camera image, using \sysname to encode the request data only needs 7 bytes, about 43.75\% of the size if we encode with the original four single precision floats.
Comparing to directly sending the raw RGB-D image (5 bytes per pixel), \sysname reduces approximately 98.3\% data usage by only needing to send on average 4249 bytes for an RGB-D image with 256x192 resolution.
This results in both less network data transfer and potentially less network time.

\subsection{Cross-frame Optimization}
\label{subsec:multi_frame_optimization}

\subsubsection{Adaptive Estimation Triggering}
\label{subsec:triggering}

Most modern camera systems provide high refresh rate, e.g. 30fps or higher.
Estimating scene lighting at the same frequency for each frame can be beneficial for achieving good visual results, but also consume significant computation and energy resources.
Additionally, it might not be necessary to update lighting information this frequently as environment lighting conditions might not change at this rate.
To avoid sending unnecessary estimation requests to the edge, we design a triggering strategy that allows \sysname client to efficiently compare lighting changes on mobile devices.

Designing effective triggering strategies involve addressing two major challenges:
\1 potential camera movements between consecutive frames;
\2 low latency requirement.
The first challenge indicates that image difference-based triggering methods are less robust as camera movements can lead to mismatched camera images between frames.
Although it is possible to leverage techniques that stitch consecutive frames, such techniques are likely to violate the latency requirement outlined in the second challenge.
As such, we design a \uspc-based triggering strategy which is less sensitive to observation point cloud changes and can be integrated as a part of our mobile GPU processing pipeline to satisfy the real-time goal.

The triggering decision making, in essence, evaluates the \uspc differences between frames and decides triggering based on the amount of difference.
Specifically, \sysname makes the triggering decision by:
\1 calculating the anchor-wise color difference (i.e., MSE of two RGB values) between two unit sphere-based point clouds that are stored in the temporary buffer and the persistent buffer, respectively; \2 obtaining the pooling averages using a sliding window of size $N$ (i.e., $N$ anchors) on the sphere for each anchor;
\3 triggering estimation when any pooled value exceeds a threshold value $\theta$.
Both the threshold and the number of nearest neighbors can be configured empirically and we set the threshold value $\theta = 0.6$ and $N$ = 4 based on our analysis in Section~\ref{subsec:triggering}.
If \sysname client decides to trigger the lighting estimation, we will continue the pipeline execution by merging the temporary buffer into the persistent buffer.
Otherwise, we will early exit the GPU pipeline execution.

\subsubsection{Providing Temporal-coherent Rendering}
\label{subsec:providing_temporal_coherent_rendering}

Lighting estimation can fall short in reflecting the physical world lighting at the exact moment.
As we described in Section~\ref{sec:fast_and_accurate} and will show in Section~\ref{sec:evaluation}, \sysname can achieve as fast as 20.67ms per lighting estimation request.
This property well positions us to use a simple yet effective approach to achieve temporal-coherent rendering.
To compensate the estimation delay, we utilize mobile ambient light sensor, which is cheap to use and provides low-latency ambient average color and intensity lighting data, to continuously adjust the rendered environment lighting per frame.
Once the \SHc response is available on the mobile device, \sysname will apply it to re-lit the virtual object.
As such, our compensation technique can improve the visual coherence when the environment lighting is changing very rapidly and handle use cases with less ideal network conditions.

To account for user movement during a single AR session, \sysname leverages mobile device's built-in accelerometer and gyroscope to obtain the camera position and orientation information.
This allows \sysname to track estimation positions (i.e., represented as world coordinates relative to the origin coordinate) and distinguish them as active and inactive positions; an inactive position is one that is outside the current camera view.
If all estimation positions are inactive, \sysname will not engage the GPU pipeline\footnote{An alternative is to leverage the triggering algorithm to proactively send \uspc to the edge for point cloud augmentation.}; as estimation positions become active, \sysname will resume its normal operations.
If \sysname client triggers the estimation and subsequently sends the \uspc corresponding to the active estimation position to the edge, \sysname server can opportunistically augment any managed \uspc at the edge.

\section{Implementation}
\label{sec:implementation}

\begin{figure}[t]
  \centering
\includegraphics[width=0.85\linewidth]{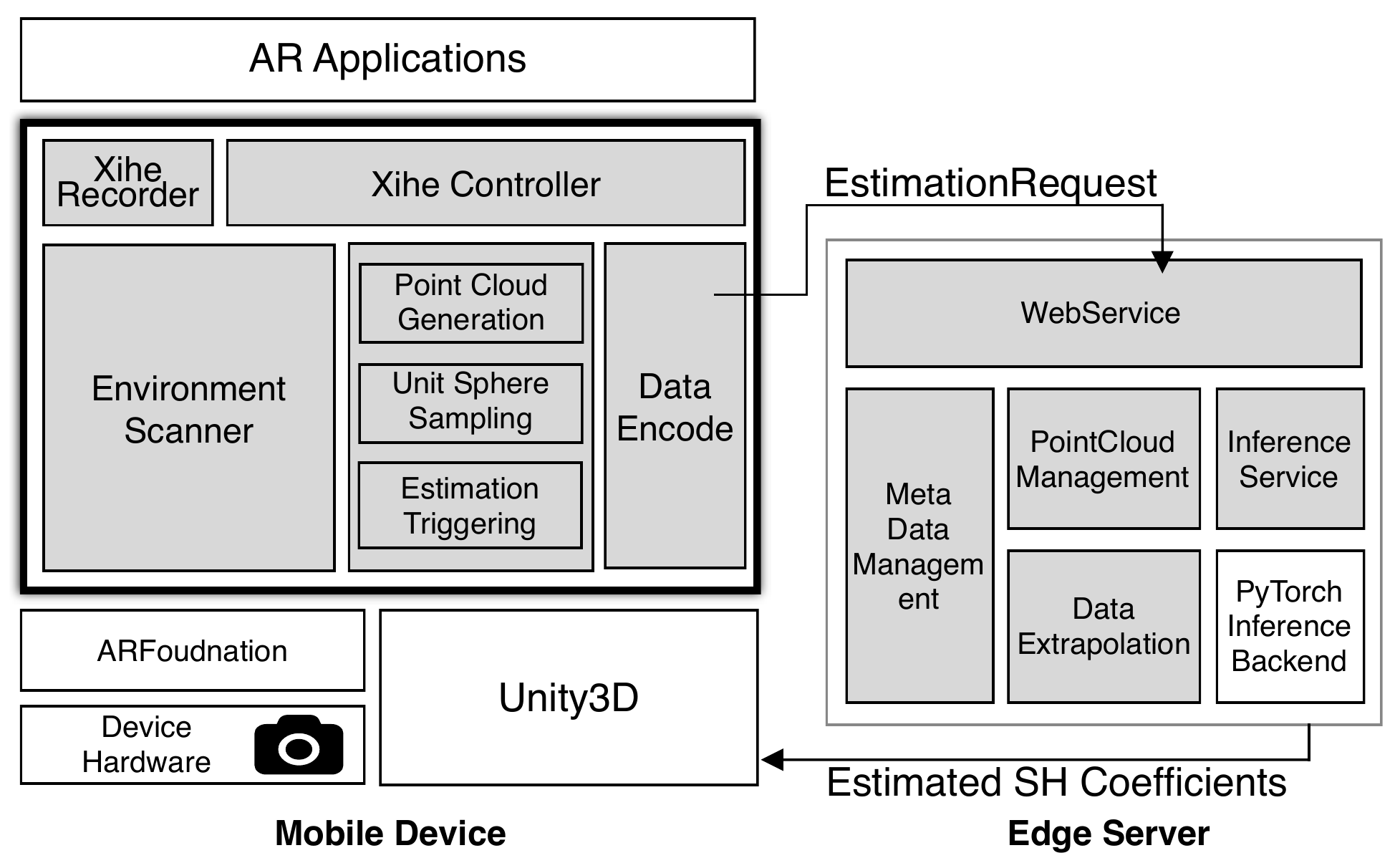}
    \vspace{-0.5em}
  \caption{\sysname architecture.
}
  \label{fig:architecture}
  \vspace{-1em}
\end{figure}

We implement \sysname in two logical components: one runs on the mobile client side and the other as a managed edge service.
Figure~\ref{fig:architecture} depicts the architecture of \sysname.
Our implementation consists of around 5K lines of code written in C\#, Python, and CUDA C++, and works with commodity hardware and software frameworks.
Specifically, \sysname client is built on top of AR Foundation 4.2.0~\cite{arfoundation} which provides basic AR functionalities and works with rendering engine including Unity3D 2020. We design \sysname client to run on a wide range of mobile hardware.
\sysname is developed as a Python-based web API server and uses PyTorch~\cite{pytorch} backend with just-in-time model optimization to host our deep learning model \xiheNet.
The server is packaged as a Docker image to facilitate deployment.

\setlength{\tabcolsep}{4pt}
\begin{table}[t]
\caption{
\sysname key classes and functions.
}
\label{tab:api}
\vspace{-0.5em}
\centering
\scriptsize{
\ra{1.3}
\begin{tabular}{lp{4.5cm}}
\toprule
\textbf{API} & \textbf{Meaning}\\
\midrule
\texttt{XiheController} & Main system controller\\
\texttt{EnvironmentScanner} & Supporting environment scan \\
\texttt{GPUDataProcessor} & Supporting point cloud processing \\
\texttt{InferenceBackend} & Provide lighting estimation inference \\
\texttt{XiheRecorder} & Provide AR session recording \\
\hline \texttt{EstimateAt(p)} & Function to estimate lighting at given position \\
\texttt{PlaceAndEstimateAt(p)} & Function to simply place virtual object and estimate placement position lighting \\
\texttt{StartRecording} & Function to start AR session recording \\
\texttt{StopRecording} & Function to stop AR session recording \\
\bottomrule
\end{tabular}
}
\vspace{-2.5em}
\end{table}

\para{Client.}
\sysname client is implemented as a C\# library that runs on Unity3D.
Table~\ref{tab:api} summarizes the provided APIs.
The mobile GPU pipeline, including \emph{Point Cloud Generation}, \emph{UnitSphere Sampling}, and \emph{Estimation Triggering} components, is implemented as Unity3D compute shaders in HLSL.
Developers can start using \sysname with both new and existing AR projects by initializing the \emph{XiheController} at the start of application life cycle.
This allows the applications to either create a new AR session or join an existing one.
Developers may ignore the implementation of other internal components or can extend \sysname to support new features, e.g., using new camera hardware through overwriting the \emph{AcquireEnvironmentScan} function in the \emph{EnvironmentScanner} class.
We provide \emph{EstimateAt} and \emph{PlaceAndEstimateAt}, two key functions to provide spatially-variant lighting estimation through the \emph{XiheController}.
The first estimation function directly returns the estimated \SHc as an ordered array, and therefore can be used in any customized rendering pipelines or shaders.
The second function works directly with Unity3D engine by supporting a simplified workflow of placing virtual objects in the format of Unity3D Prefabs into the physical surrounding.
If sufficient lighting change is detected, both functions will trigger HTTP requests to send the encoded \uspc to the server.
\sysname also has a built-in AR session recorder that captures the essential AR session information, including lighting estimation position, RGB-D camera feed, camera pose and ambient light sensor data. This recorder is built on top of the same \texttt{EnvironmentScanner} used in \sysname's estimation workflow. We provide two APIs, \texttt{StartRecording} and \texttt{StopRecording}, to control the recording process.

\para{Server.}
\sysname server is built on top the \emph{Tornado} web framework and the PyTorch inference backend with just-in-time model optimization.
\sysname server provides two key features, namely the \emph{AR session management} and the \emph{lighting estimation request processing}.
The AR session management service is provided by three components, i.e., \emph{MetaData Management}, \emph{PointCloud Management}, and \emph{Data Extrapolation}.
Specifically, the \emph{MetaData Management} is responsible for storing AR session's basic information, such as unique session ID issued by the server, client-specific identifier, and timestamp.
To connect to the \sysname server, each \sysname client will either create a new AR session by posting requests or join an existing AR session by providing the session ID.
Additionally, we use the \emph{NumPy} library to perform tensor operations on HTTP payload in the byte format to achieve historical data extrapolation.
A pretrained \xiheNet (number of anchors = 1280) is included and managed by the PyTorch inference backend.
\sysname server is packaged as a Docker image and can be setup with minimal configuration effort.

\para{A reference AR application.}
We include an example AR application implemented with \sysname APIs and the ARKit V4.0 plugin from the ARFoundation. The resulting application can be compiled to run on iOS and macOS.
This reference application allows a mobile user to place 3D virtual objects by selecting rendering positions in the current camera view.
For each virtual object, the mobile application will generate one logical lighting estimation request per frame.
Depending on the detected lighting conditions, \sysname client will trigger one or more physical lighting estimation requests which will send the encoded \uspc to the \sysname server for inference.
The number of physical lighting estimation requests per frame is by default depending on the object size but can also be configured by the mobile AR developers for performance and quality trade-offs.
The 3D objects will be rendered with spatially-variant lighting information provided by \sysname.
Lastly, users can easily record the AR session with \sysname's session recorder, facilitating real-world record-and-replay experiments.

\section{Evaluation}
\label{sec:evaluation}

We conducted our experiments by using an example AR application which uses \sysname's APIs for obtaining spatially-variant lighting estimation.
We used three different devices, a MacBook Pro 15'', a 2nd generation iPad Pro 11'' with a built-in Lidar sensor, and an iPhone 11 Pro, to evaluate the on-device performance.
Our edge service is on a desktop running Ubuntu 20.04 with a 16 core AMD Ryzen Threadripper 2950X CPU, 128GB memory, and a Nvidia RTX 2080Ti GPU.
We quantified \sysname's performance in terms of end-to-end lighting estimation time, accuracy, and visual effects and compared it to the commercial baseline ARKit~\cite{arkit}, an academic framework GLEAM~\cite{prakash2019gleam}, and a 3D vision estimation pipeline~\cite{pointar_eccv2020} where appropriate.
\sysname can deliver spatially-variant lighting estimation as fast as 20.67ms and achieves visually-coherent rendering effects.
We further evaluate how each proposed technique and configuration contributes to \sysname's performance with a detailed breakdown study, e.g., with different sampling strategy, anchor size, and mobile network condition, and a lab-based real-world evaluation.

\subsection{End-to-end Performance}

\begin{figure}[t]
\includegraphics[width=0.95\linewidth]{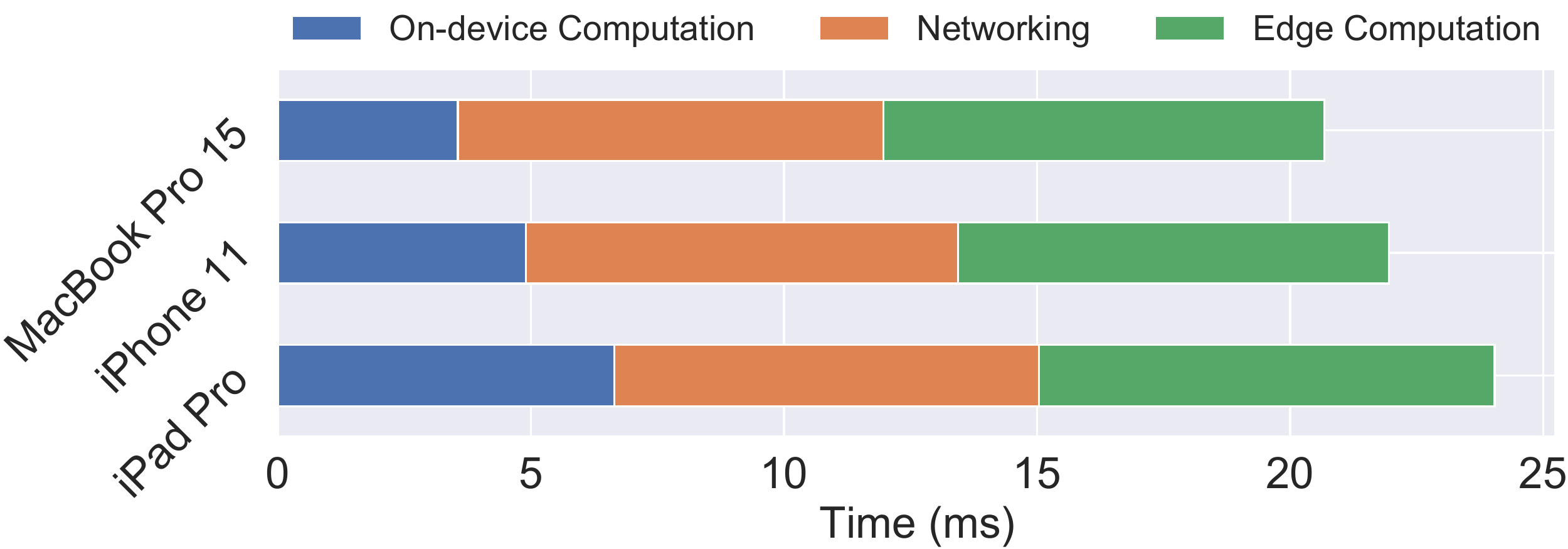}
  \vspace{-0.5em}
  \caption{\sysname end-to-end time.
  \textnormal{
  \sysname lighting estimation via the university WiFi, can complete in as fast as 20.67ms.
}
}
  \label{fig:end_to_end_time}
  \vspace{-1.5em}
\end{figure}

\begin{figure}[t]

\centering
    \includegraphics[width=\linewidth]{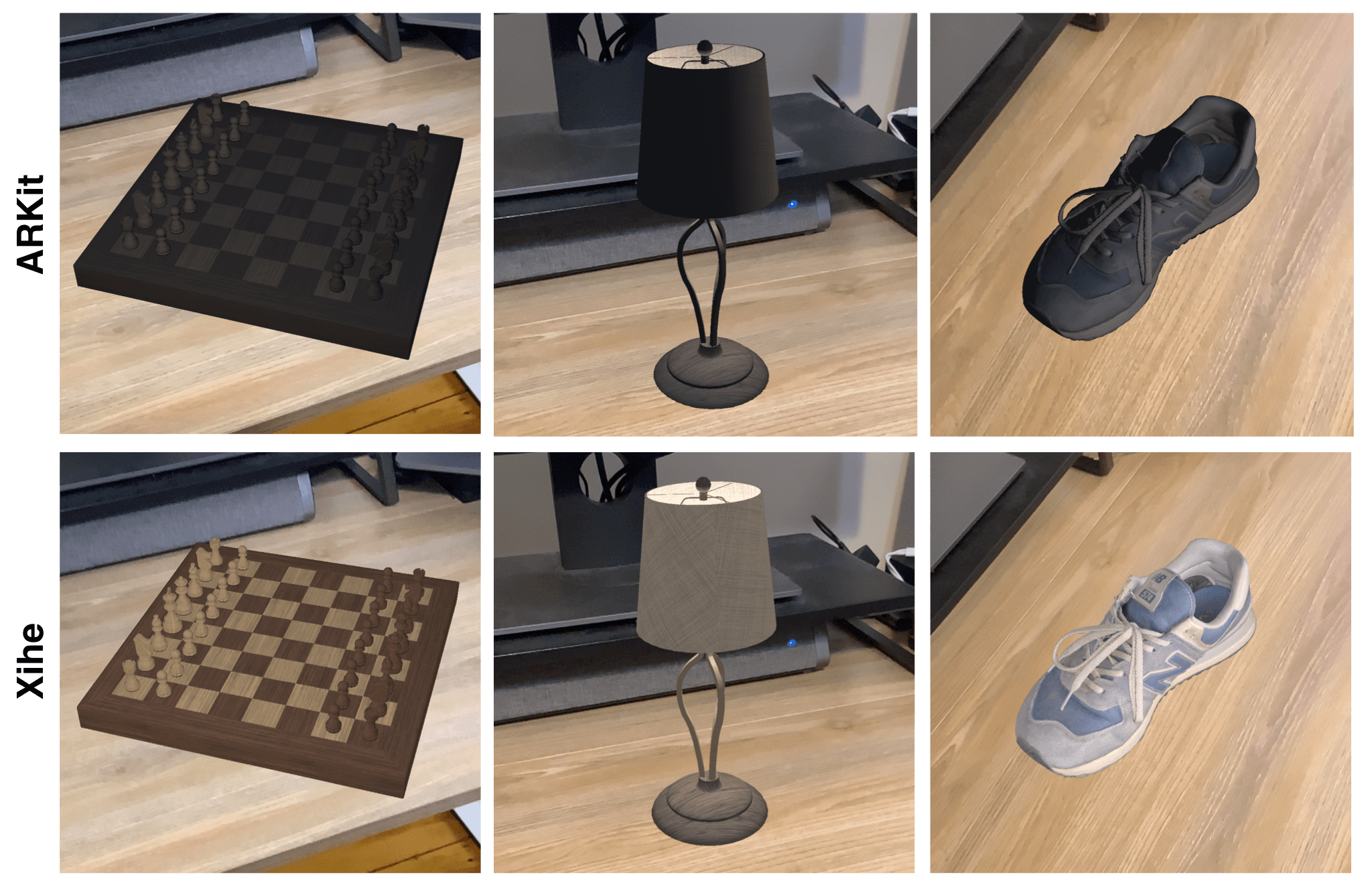}

    \vspace{-0.5em}

    \caption{
AR scenes rendered with lighting information provided by ARKit and our framework \sysname.
    }
    \vspace{-1.5em}
    \label{fig:eval_rendering}
\end{figure}

We demonstrate the end-to-end performance achieved by \sysname; \sysname takes less than 24.04ms to complete in all three devices with the university WiFi as shown in Figure~\ref{fig:end_to_end_time}.
As such, \sysname not only can support 30fps refresh rate but also takes 19.9\% less time than GLEAM~\cite{prakash2019gleam}.
The on-device GPU computation takes less than 6.65ms to finish running on all three devices.
This indicates that \sysname can support a wide range of mobile devices.
Figure~\ref{fig:eval_rendering} compares the visual effects of three 3D objects. Given the same virtual object and the same environment lighting condition, \sysname's reference AR application is able to display the virtual object in a photorealistic manner.
However, when using the ARKit's ambient lighting estimation APIs, objects will be rendered with less desirable effect as only ambient lighting intensity and color information are available.

\subsection{Performance Breakdown}

\setlength{\tabcolsep}{4pt}
\begin{table}[t]
\centering

\caption{
Mobile computation breakdown.
\textnormal{All components except the first and the last run in the GPU pipeline.}
}

\vspace{-1em}

\scriptsize{
\begin{tabular}{lrrr}
\toprule
\begin{tabular}[l]{@{}l@{}}\textbf{On-device}\\ \textbf{Component}\end{tabular} &
\begin{tabular}[r]{@{}r@{}}\textbf{MacBook Pro}\\ \textbf{Avg. Time} (ms)\end{tabular} &
\begin{tabular}[r]{@{}r@{}}\textbf{iPhone}\\ \textbf{Avg. Time} (ms)\end{tabular}
 & \begin{tabular}[r]{@{}r@{}}\textbf{iPad Pro}\\ \textbf{Avg. Time} (ms)\end{tabular}\\
\midrule
\texttt{AcquireEnvironmentScan} & N/A & N/A & 1.458 \\
\texttt{GeneratePointCloud} & 0.063 & 0.193 & 0.221 \\
\texttt{UniSphereSampling} & 0.015 & 0.061 & 0.063 \\
\texttt{MergeTemporaryBuffer} & 0.007 & 0.024 & 0.029 \\
\texttt{MakeTriggeringDecision} & 1.34 & 2.185 & 3.040 \\
\texttt{MergePersistentBuffer} & 0.013 & 0.057 & 0.062 \\
\texttt{EncodeBuffer} & 2.16 & 2.332 & 2.832 \\
\bottomrule
\end{tabular}
}
\label{tab:profile_gpu_pipeline}
\vspace{-0.5em}
\end{table}
\setlength{\tabcolsep}{1.4pt}

\subsubsection{On-device performance}
To quantify the performance of \sysname client, we measure the time breakdown of each component that runs on the mobile device.
Table~\ref{tab:profile_gpu_pipeline} shows the average time across five runs measured with the Unity3D built-in profiling tool~\cite{unity_profiler}.
For the MacBook Pro and iPhone measurement, since they do not have built-in Lidar sensors, we randomly selected five test data.
First, we observe that the total on-device time excluding the \emph{AcquireEnvironmentScan} step using MacBook Pro, iPhone and iPad Pro are
3.57ms, 4.90ms and 6.65ms, respectively.
This is expected as the results matches the devices' GPU computation capabilities.
Second, the long GPU time of two dominating components (i.e., \texttt{MakeTriggeringDecision} and \texttt{EncodeBuffer}) are likely due to the callback functions needed for communicating between CPU and GPU and non-continuous memory access during encoding.

\setlength{\tabcolsep}{4pt}
\begin{table}[t]
\centering
\vspace{-0.5em}
\caption{
Edge computation breakdown.
}
\vspace{-1em}
\label{tab:profile_edge}
\scriptsize{
\begin{tabular}{lrrr}
\toprule
\begin{tabular}[l]{@{}l@{}}\textbf{Edge}\\ \textbf{Component}\end{tabular} & \begin{tabular}[r]{@{}r@{}}\textbf{512 Anchors}\\ (ms)\end{tabular} &
\begin{tabular}[r]{@{}r@{}}\textbf{1280 Anchors}\\ (ms)\end{tabular} &
\begin{tabular}[r]{@{}r@{}}\textbf{2048 Anchors}\\ (ms)\end{tabular}\\
\midrule
\texttt{Decoding} & 0.23 & 0.30 & 0.32 \\
\texttt{Extrapolation} & 0.01 & 0.01 & 0.01 \\
\texttt{Inference} & 3.99 & 5.91 & 10.80 \\
\bottomrule
\end{tabular}
}
\vspace{-1.5em}
\end{table}
\setlength{\tabcolsep}{1.4pt}

\subsubsection{Edge performance}
To quantify the performance of \sysname server, we measure the time breakdown of each component that runs on a GPU-equipped desktop.
Table~\ref{tab:profile_edge} shows the average time across the entire test dataset.
First, we observe that both the \texttt{decoding} and \texttt{inference} time increase with the number of anchors. This is expected as \uspc with more anchors are likely to have more encoded points that need to be decoded and transformed.
Second, the point cloud extrapolation only takes 0.01ms but provides the opportunity to improve the lighting estimation accuracy, i.e., by boosting the input point cloud's entropy with historical data.

\subsubsection{Network performance}

\begin{figure}[t]
    \centering
    \begin{subfigure}[b]{0.45\linewidth}
        \centering
\includegraphics[width=\linewidth]{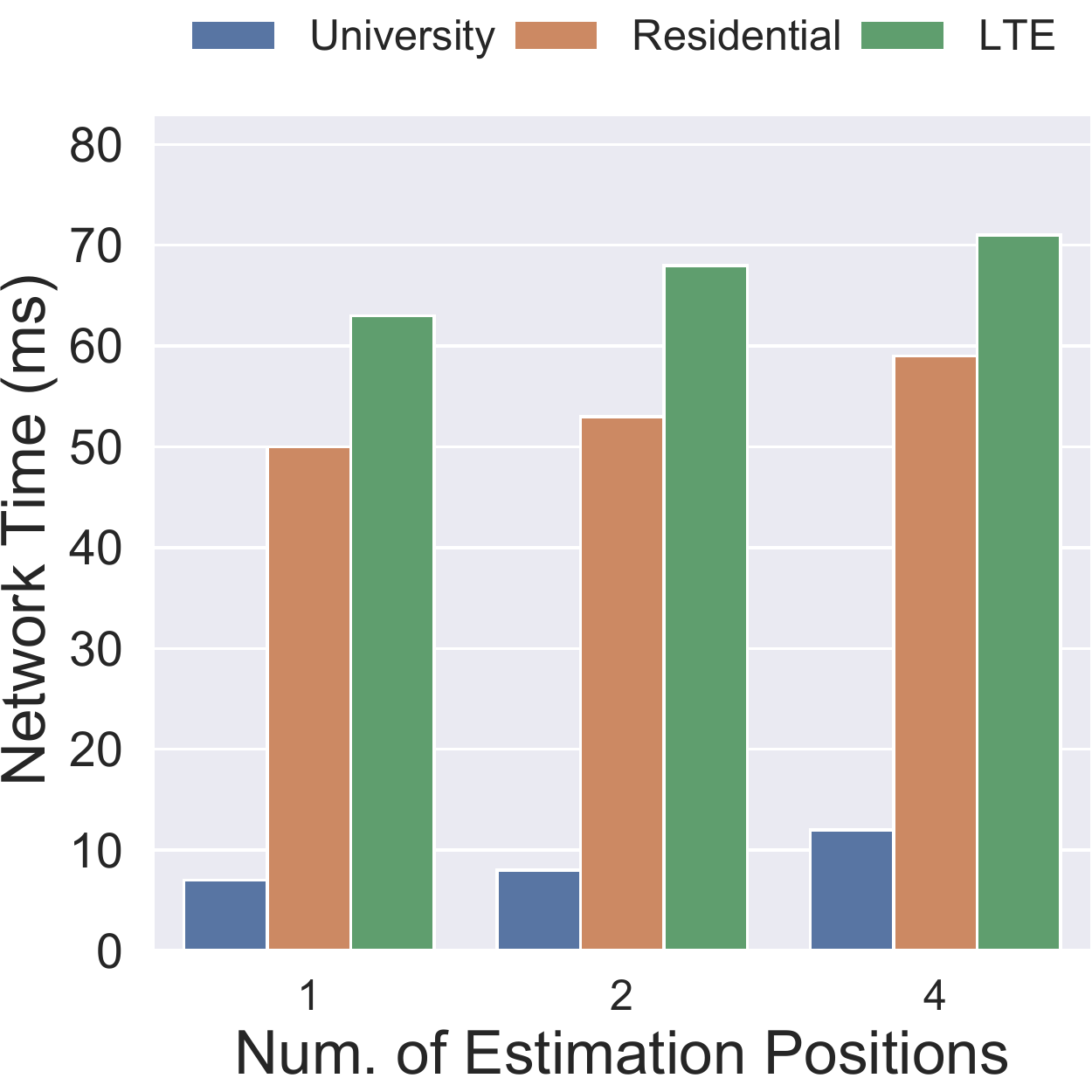}
        \caption{iPhone.}
    \end{subfigure}\quad
    \begin{subfigure}[b]{0.45\linewidth}
        \centering
\includegraphics[width=\linewidth]{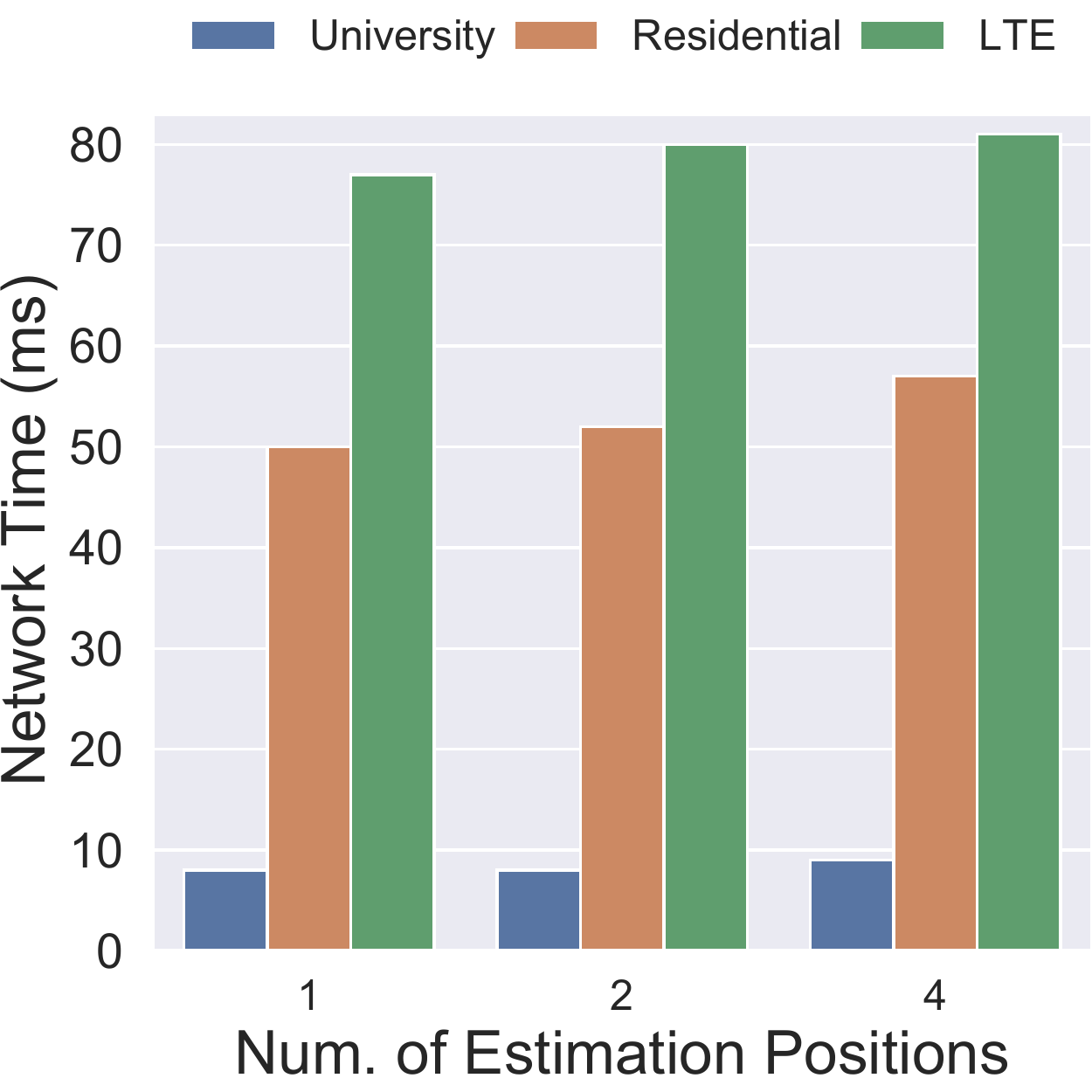}
        \caption{iPad Pro (LTE via hotspot).}
    \end{subfigure}
\vspace{-1em}
    \caption{\sysname network time.
    \textnormal{We measure the time needed to transfer the encoded point cloud of 1280 anchors and to receive the lighting estimation \SHc.
}
}
    \label{fig:xihe_network_time}
    \vspace{-1em}
\end{figure}

Figure~\ref{fig:xihe_network_time} shows the network time under different network conditions and user interactions.
If the user places larger objects, i.e., more positions to estimate, the network time increases sublinearly under all network conditions.
For example, for iPad Pro that uses the university WiFi, the network time to handle four estimation positions is about 1.28X that of one estimation position.
Both residential WiFi and LTE take several times longer than using the university WiFi, indicating the need to properly deploy the \sysname server.
Even under undesirable network condition, e.g., iPhone with LTE, \sysname can still generate one lighting estimation in about 79.2ms which is lower than the 400ms needed by GLEAM to generate high-fidelity estimation~\cite{prakash2019gleam}.

\subsection{Impact of Point Cloud Sampling}
\label{subsec:impact_pcs}

\begin{figure}[t]
    \centering
\includegraphics[width=0.95\linewidth]{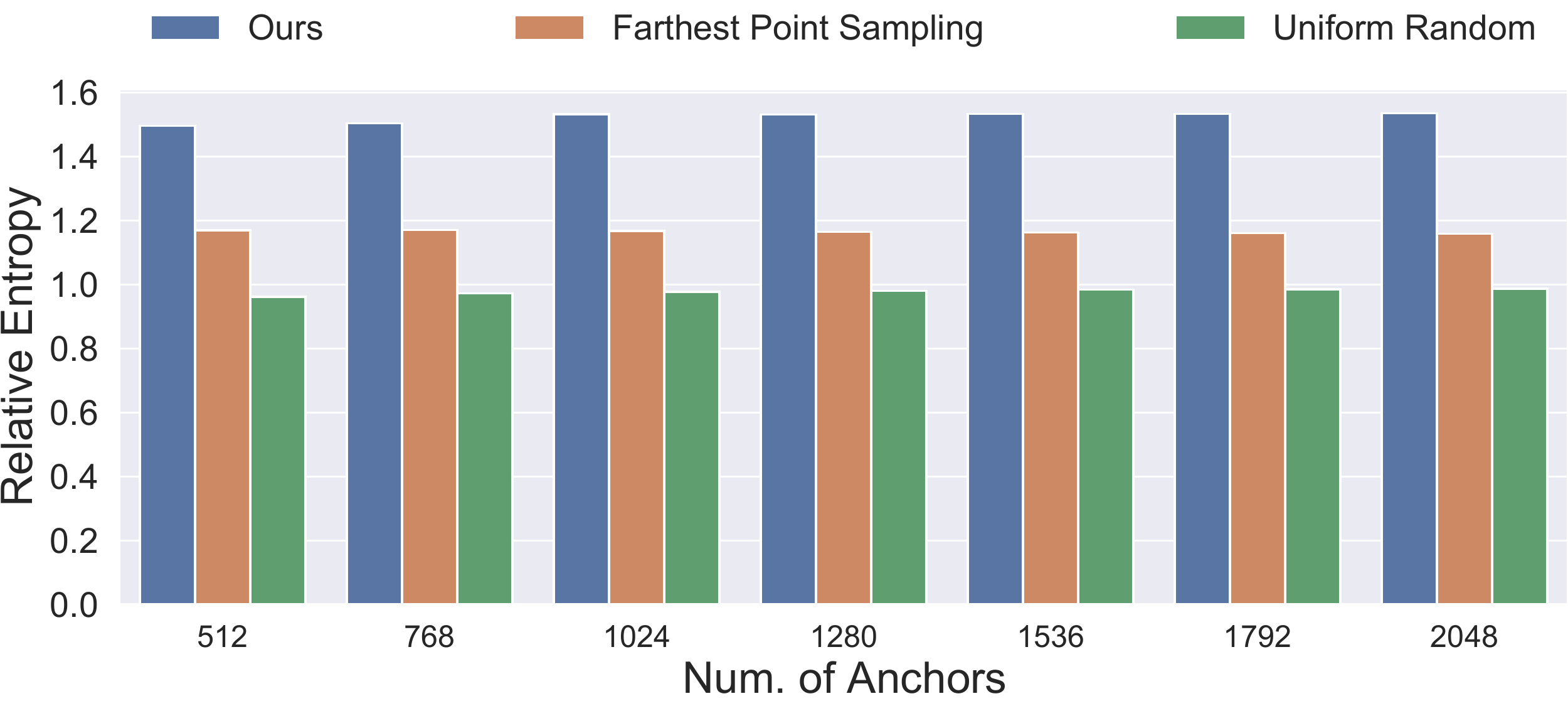}
\vspace{-1em}
    \caption{
Relative entropy comparison of different point cloud sampling techniques.
\textnormal{Our \uspcs achieves 55.74\% and 30.80\% better entropy compared to uniform random sampling~\cite{pointar_eccv2020} and farthest point sampling techniques~\cite{pointnet_plus_plus_nips}, respectively.}
}
    \label{fig:eval_pcs}
    \vspace{-1.5em}
\end{figure}

\para{Entropy comparison.}
We use the point cloud test dataset from PointAR~\cite{pointar_eccv2020} and generate three variants using the \emph{uniform random sampling}~\cite{pointar_eccv2020}, \emph{farthest point sampling}~\cite{pointnet_plus_plus_nips}, and our proposed \uspcs techniques.
For each point cloud (and its downsampled versions), we first calculate the entropy using Equation~\eqref{eq:entropy} and divide it by the raw point cloud entropy.
Figure~\ref{fig:eval_pcs} compares the relative entropy.
First, our \uspcs approach is more effective in preserving the entropy, with on average 0.545 higher relative entropy than using the uniform random sampling, and 0.359 higher than using the farthest point sampling.
Second, using more anchors can improve the entropy but the improvement plateaus after 1280.
This observation suggests that using 1280 anchors can be effective.
Later in Section~\ref{subsec:eval_3d_estimator} we will compare and show that \uspc technique also achieves better estimation accuracy.

\begin{figure}[t]
    \centering
    \begin{subfigure}[b]{0.45\linewidth}
        \centering
\includegraphics[width=\linewidth]{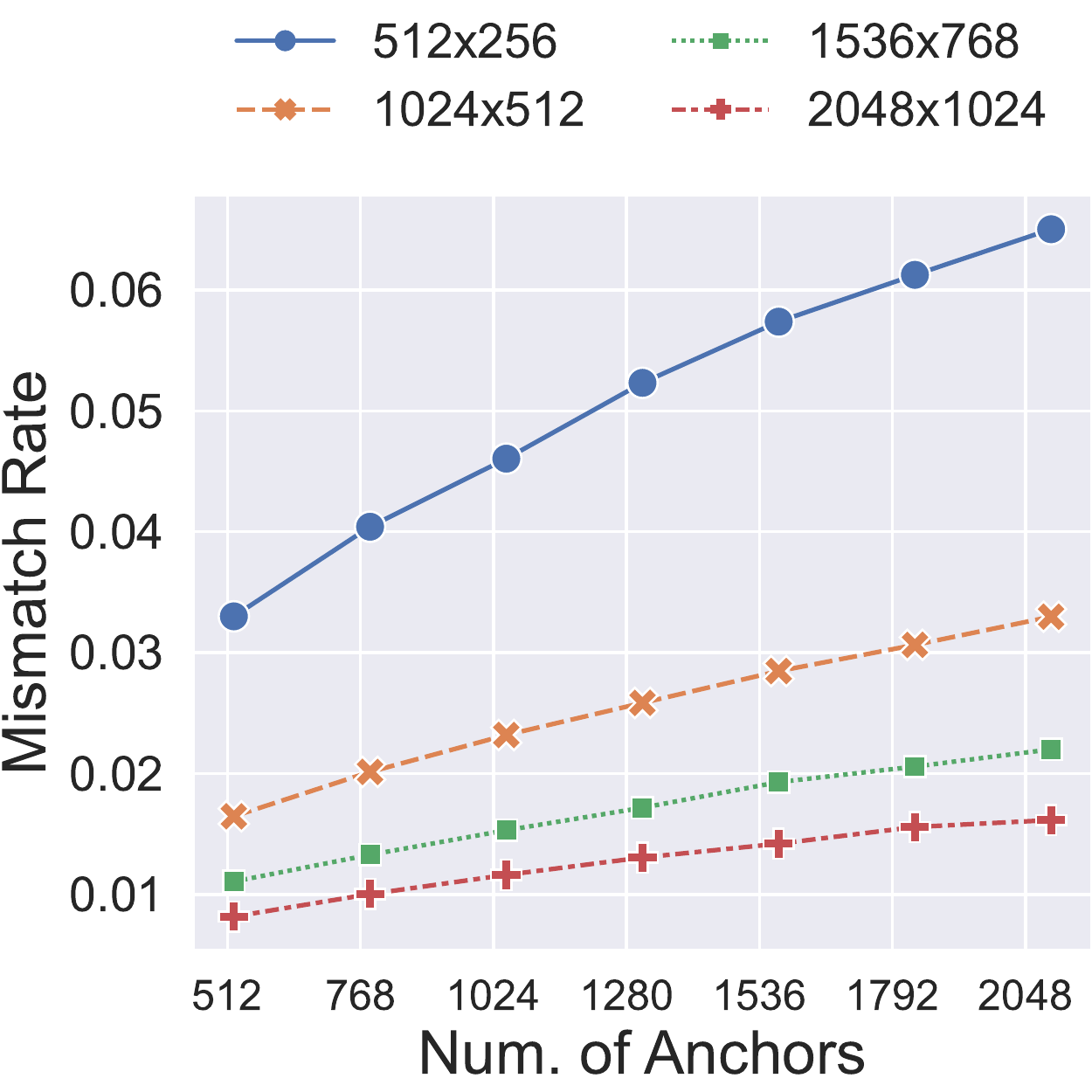}
        \caption{Acceleration mismatch}
        \label{subfig:acceleration_mismatch}
    \end{subfigure}
    \quad
    \begin{subfigure}[b]{0.45\linewidth}
        \centering
\includegraphics[width=\linewidth]{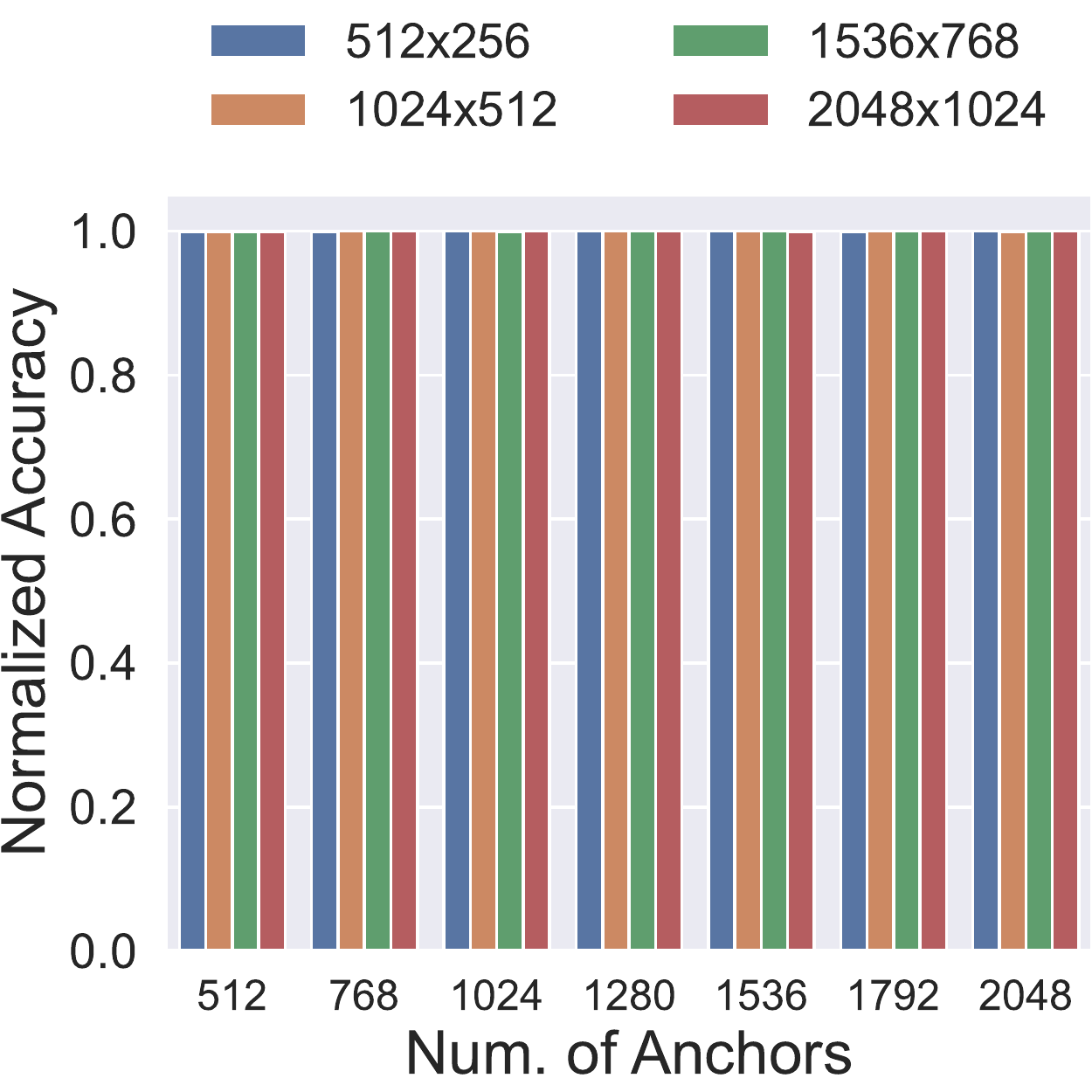}
        \caption{Impact on estimation accuracy}
        \label{subfig:acc_estimation_accuracy}
    \end{subfigure}
    \vspace{-1em}
    \caption{
Impact of acceleration grids.
        \textnormal{Even though acceleration grids might erroneously match a small percent of points, it has minimal impact on the estimation accuracy.}
}
    \label{fig:acc_mismatch}
    \vspace{-.5em}
\end{figure}

\para{Impact of acceleration grids.}
In this section, we analyze the error associated with the acceleration grid with the \emph{mismatch rate} metric, calculated by comparing the colored anchors with and without acceleration.
We first randomly generate a set of 1M points in a cubic 3D space with an edge length of 10 meters to simulate common AR application scenarios in real-world environments.
Figure~\ref{subfig:acceleration_mismatch} shows the mismatch rate using different acceleration grid sizes.
First, as the grid size increases (i.e., corresponding to pre-sampled more points), the mismatch rate decreases.
For example, with an acceleration grid of 1024x512 and an anchor size of 1280, we observe that 97.36\% of points were matched to the same anchor without using acceleration.
Second, for a given grid size, the mismatch rate also depends on and grows with the number of anchors.
However, the impact is relatively small compared to the choice of acceleration grid size and is within 1\% for the range of anchor numbers.

We further investigate the impact of our acceleration mechanism on the lighting estimation accuracy.
Figure~\ref{subfig:acc_estimation_accuracy} shows the normalized accuracy, calculated by comparing the accuracy achieved using acceleration and the ground truth accuracy using \xiheNet.
We see that neither the acceleration grid size and the anchor number impact the lighting estimation accuracy.
This also suggests that our \xiheNet has a good generalization.

\subsection{Performance of Encoding}

\begin{figure}[t]
    \centering
    \begin{subfigure}[b]{0.48\linewidth}
        \centering
\includegraphics[width=\linewidth]{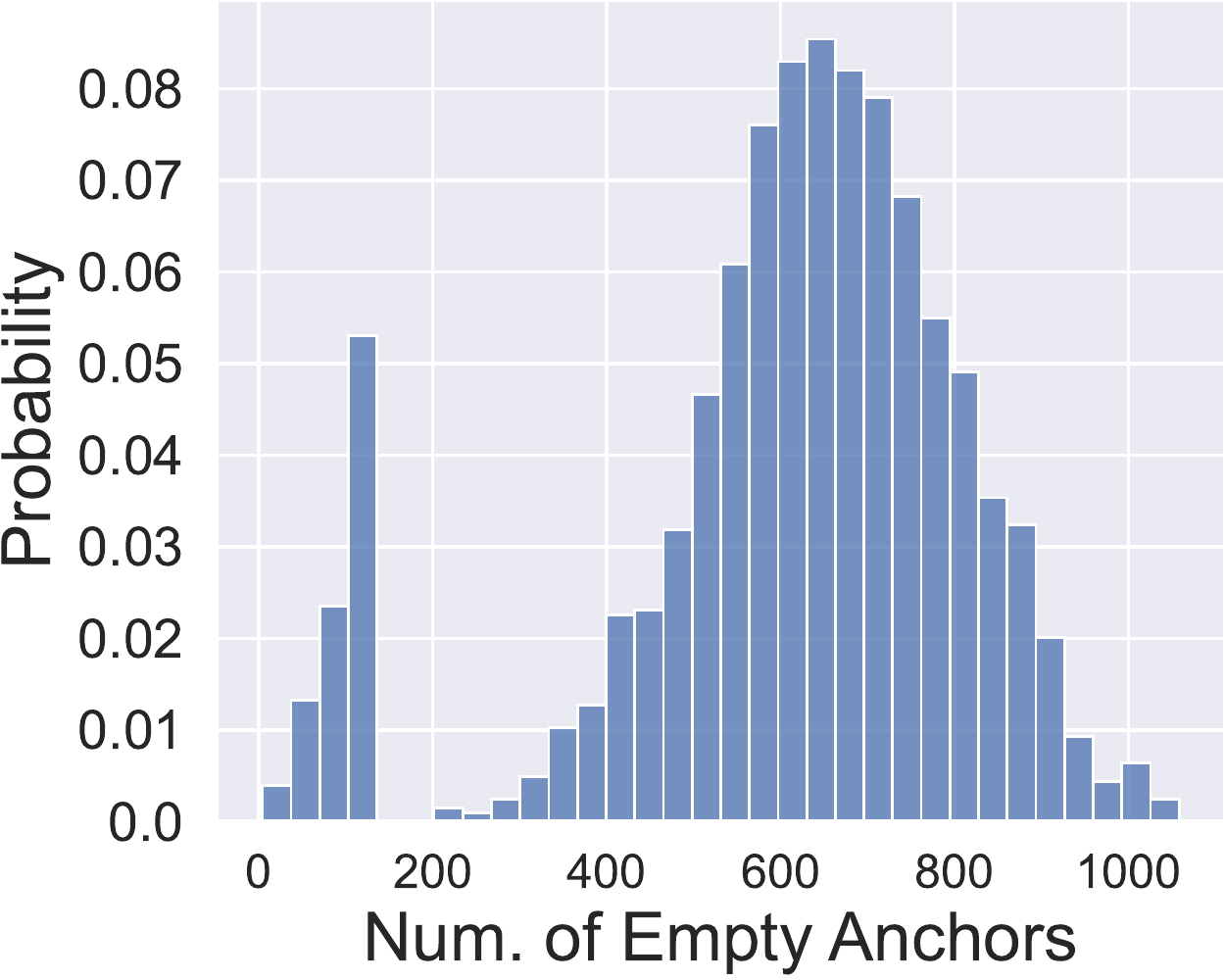}
        \caption{Number of Anchors = 1280.}
        \label{fig:empty_anchors_distribution}
    \end{subfigure}
\begin{subfigure}[b]{0.48\linewidth}
        \centering
\includegraphics[width=\linewidth]{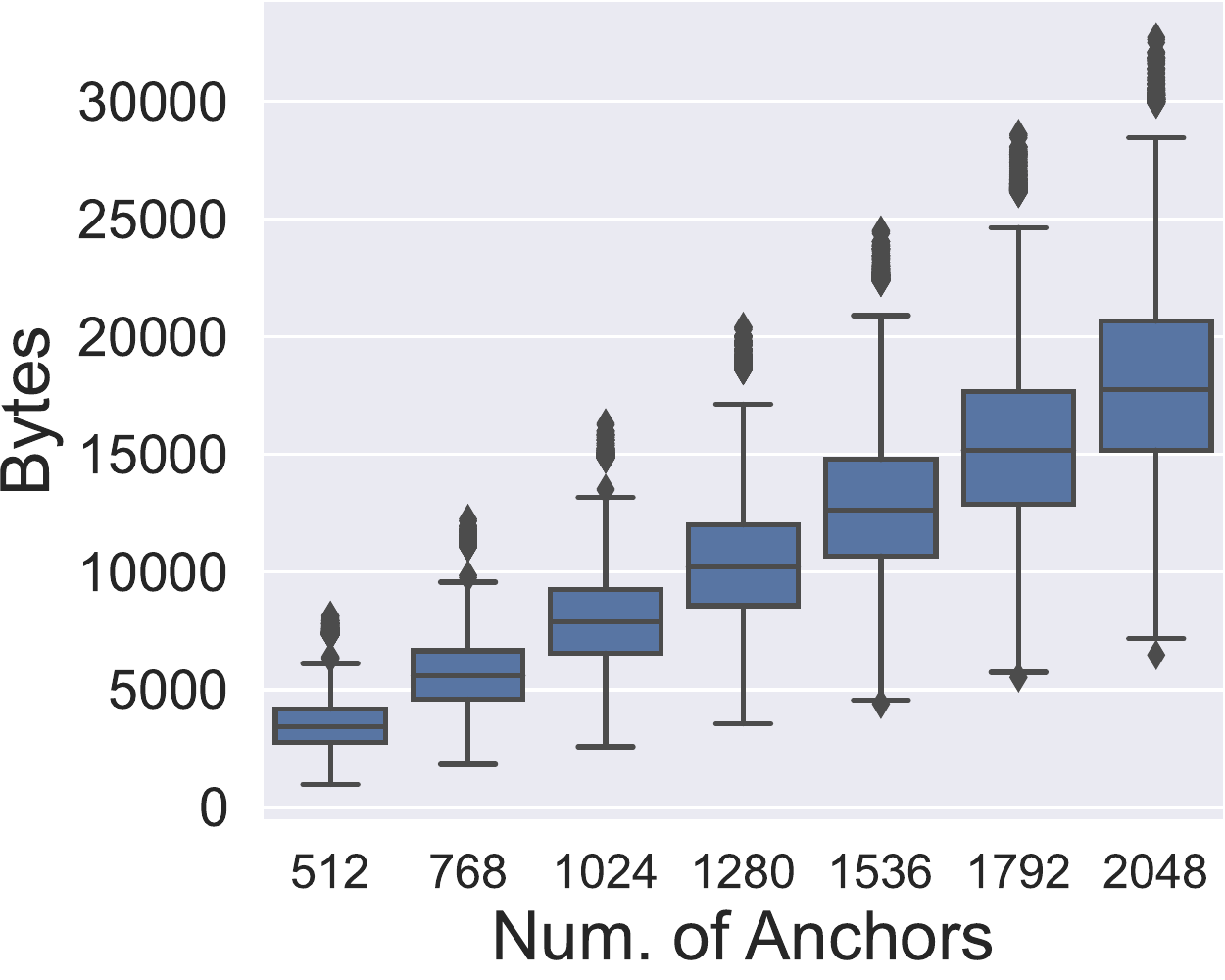}
        \caption{Non-empty anchors.}
        \label{fig:non_empty_anchors}
    \end{subfigure}
    \vspace{-1em}
    \caption{
        Empirical characterization of \uspc generated from the test dataset~\cite{pointar_eccv2020}.
}
    \vspace{-1em}
    \label{fig:network_emperical}
\end{figure}

\setlength{\tabcolsep}{4pt}
\begin{table}[t]
\centering
\caption{
Comparison of different encoding methods.
}
\vspace{-0.5em}
\scriptsize{
\begin{tabular}{lrrr}
\toprule
\textbf{Encoding Method} & \begin{tabular}[r]{@{}r@{}}\textbf{Unit Size}\\ (bytes)\end{tabular}
& \begin{tabular}[r]{@{}r@{}}\textbf{Avg. Size}\\ (bytes)\end{tabular}
& \begin{tabular}[r]{@{}r@{}}\textbf{Avg. Time}\\ (ms)\end{tabular} \\
\midrule
\texttt{float32} & 16 & 20480 & 0.003 \\
\texttt{float32} + striping & 18 & 10926 & 1.793  \\
\texttt{uint8} + \texttt{float16} & 5  & 8960 & 1.675 \\
\texttt{uint8} + \texttt{float16} + striping (Ours) & 7 & 4249 & 1.003 \\
\bottomrule
\end{tabular}
}
\label{tab:network_bytes}
\vspace{-1em}
\end{table}

\begin{figure}[t]
    \centering
    \begin{subfigure}[b]{0.45\linewidth}
        \centering
\includegraphics[width=\linewidth]{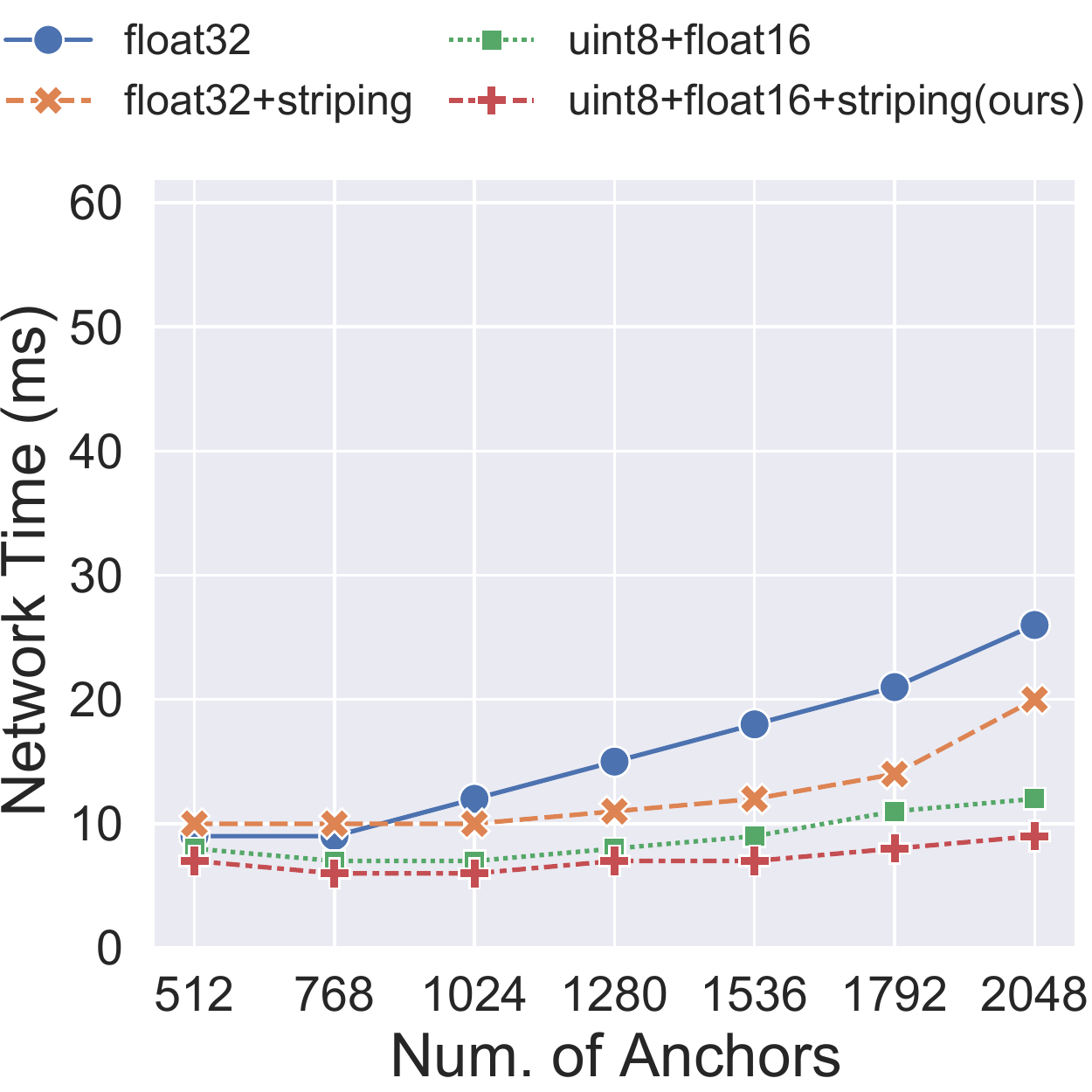}
        \caption{One estimation position.}
    \end{subfigure}\quad
    \begin{subfigure}[b]{0.45\linewidth}
        \centering
\includegraphics[width=\linewidth]{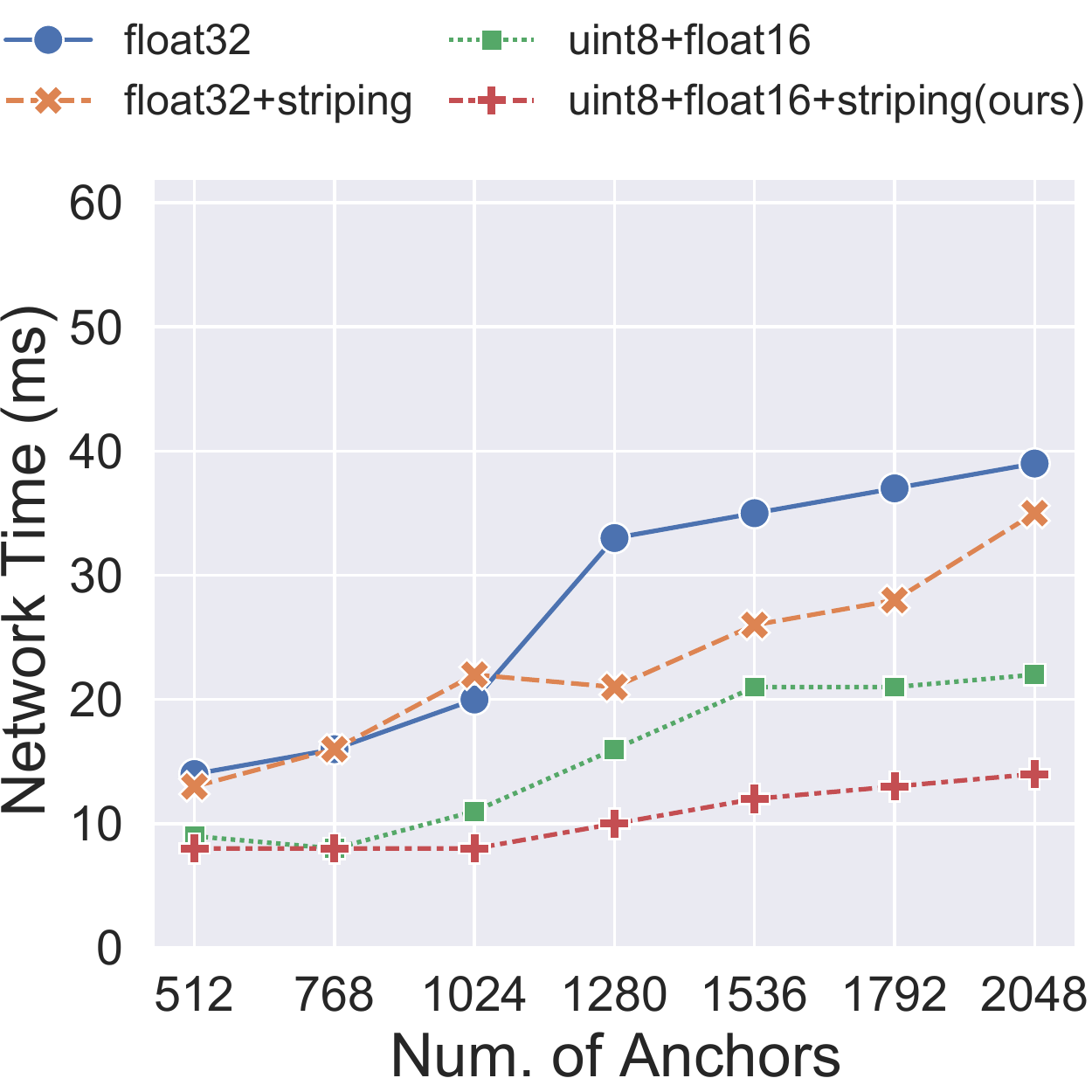}
        \caption{Two estimation positions.}
    \end{subfigure}
    \quad
    \begin{subfigure}[b]{0.45\linewidth}
        \centering
\includegraphics[width=\linewidth]{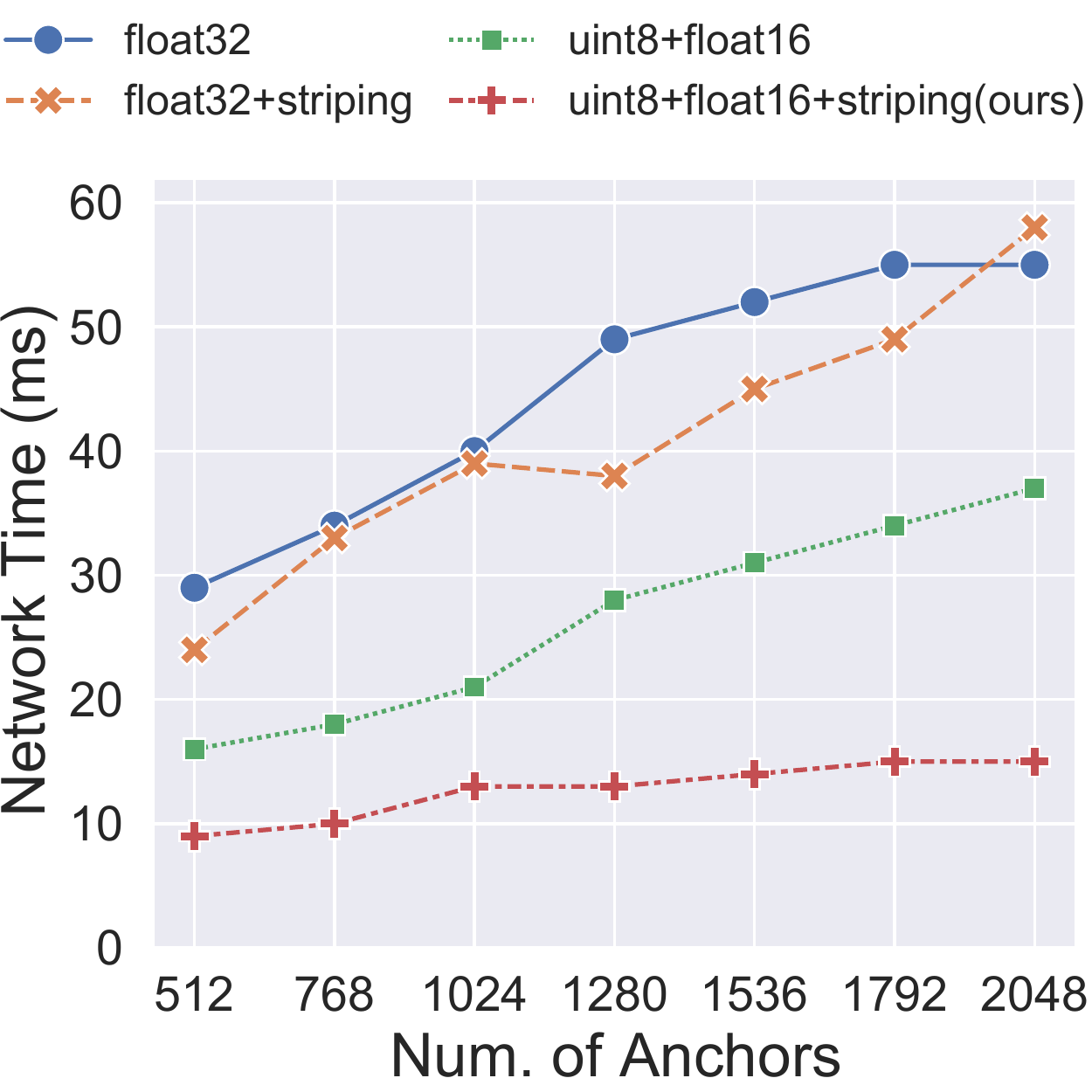}
        \caption{Four estimation positions.}
    \end{subfigure}
    \quad
    \begin{subfigure}[b]{0.45\linewidth}
        \centering
\includegraphics[width=\linewidth]{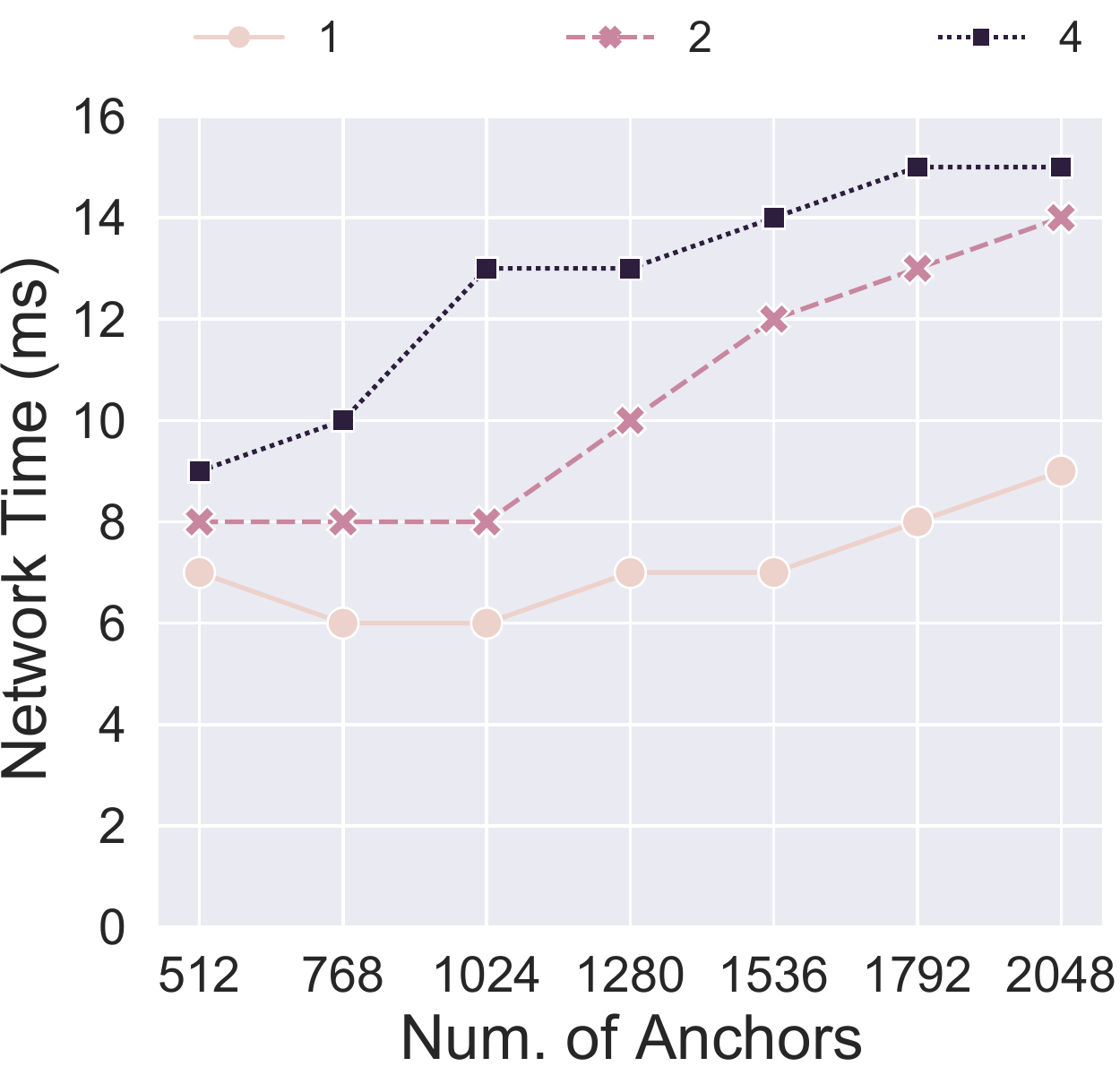}
        \caption{Our practical encoding.}
    \end{subfigure}
\vspace{-1em}
    \caption{
    Network performance of different encoding methods.
    \textnormal{The mobile AR application connects to the \sysname server via the university WiFi.}
}
    \label{fig:encoding_network_time}
    \vspace{-1.5em}
\end{figure}

Next, we evaluate the effectiveness of different encoding methods.
Figure~\ref{fig:network_emperical} presents the empirical characterization by performing the \uspcs technique to the raw point clouds from the test dataset.
We find that \uspc generated from a single RGB-D camera image usually contains many uninitialized anchors (i.e., empty anchors), as shown in Figure~\ref{fig:empty_anchors_distribution}.
For example, when setting the number of anchors to be 1280, we observe that more than half anchors are empty.
Figure~\ref{fig:non_empty_anchors} shows the required bytes for encoding only non-empty anchors using \texttt{float32}.

Table~\ref{tab:network_bytes} compares the required bytes and time to encode our \uspc using different methods.
Even though directly encoding using \texttt{uint8} for RGB values and \texttt{float16} for depth information only requires 5 bytes per anchor, it takes more than twice as many bytes to transfer the entire \uspc than our encoding approach.
Further, as the striping operation is cheaper, taking about 0.09ms on the iPad Pro, our encoding method improves the total encoding time by 1.67X compared to directly encoding with \texttt{unit8+float16}.

Figure~\ref{fig:encoding_network_time} shows the median network time to transfer the \uspc from the MacBook Pro to the \sysname server via the university WiFi.
Results for other devices and network conditions (residential WiFi and T-mobile LTE) exhibit similar trends and are omitted.
First, we observe that our encoding method takes as little as 26.3\% network time compared to three baselines under all combinations of anchor numbers and estimation positions.
Second, the time taken by our encoding method grows slower with the number of anchors compared to other encoding approaches.

\subsection{3D Vision-based Estimator Performance}
\label{subsec:eval_3d_estimator}

\begin{figure}[t]
    \centering
    \begin{subfigure}[b]{0.45\linewidth}
        \centering
\includegraphics[width=\linewidth]{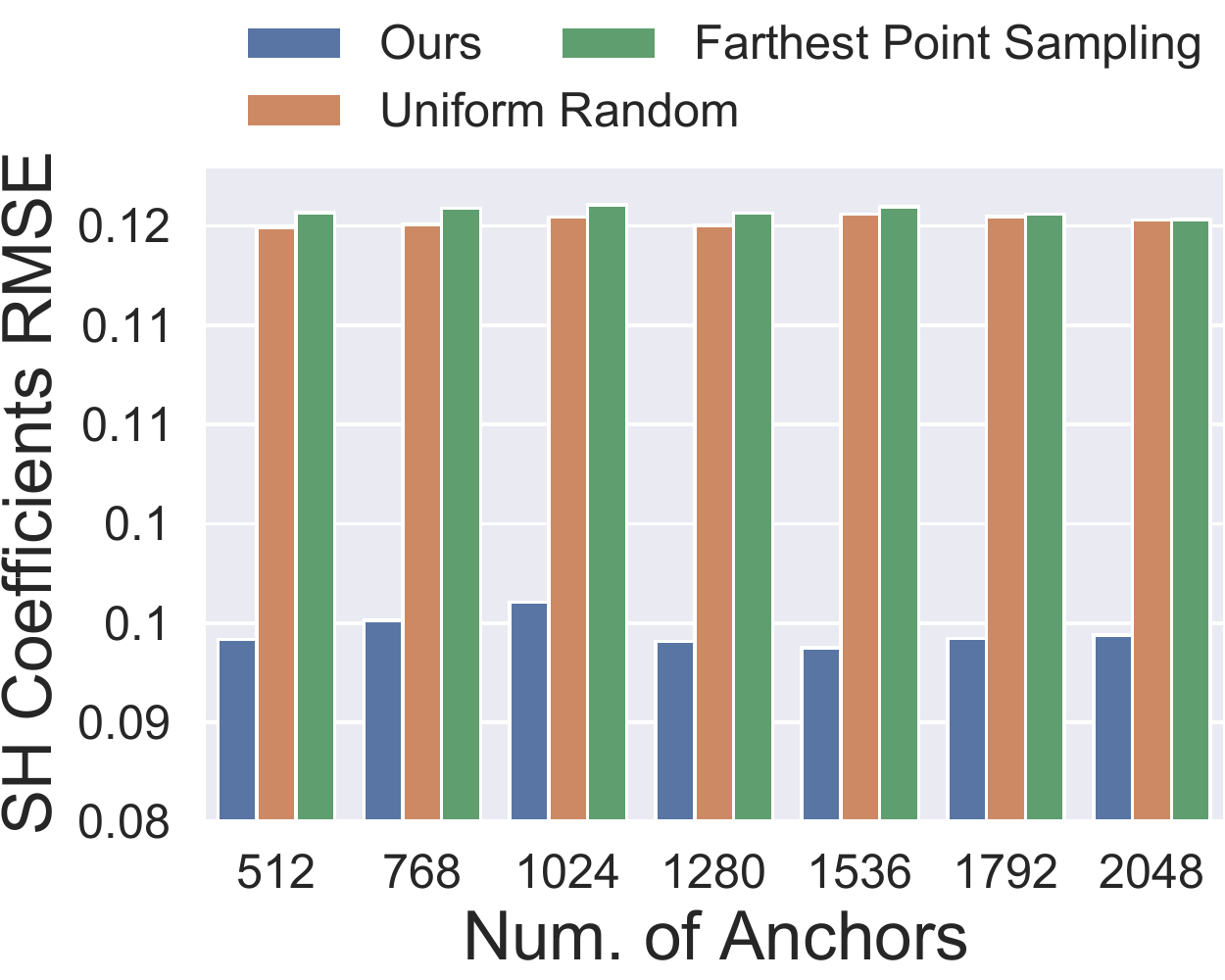}
        \caption{Lighting estimation error.
}
        \label{subfig:xihenet_accuracy}
    \end{subfigure}\quad
    \begin{subfigure}[b]{0.45\linewidth}
        \centering
\includegraphics[width=\linewidth]{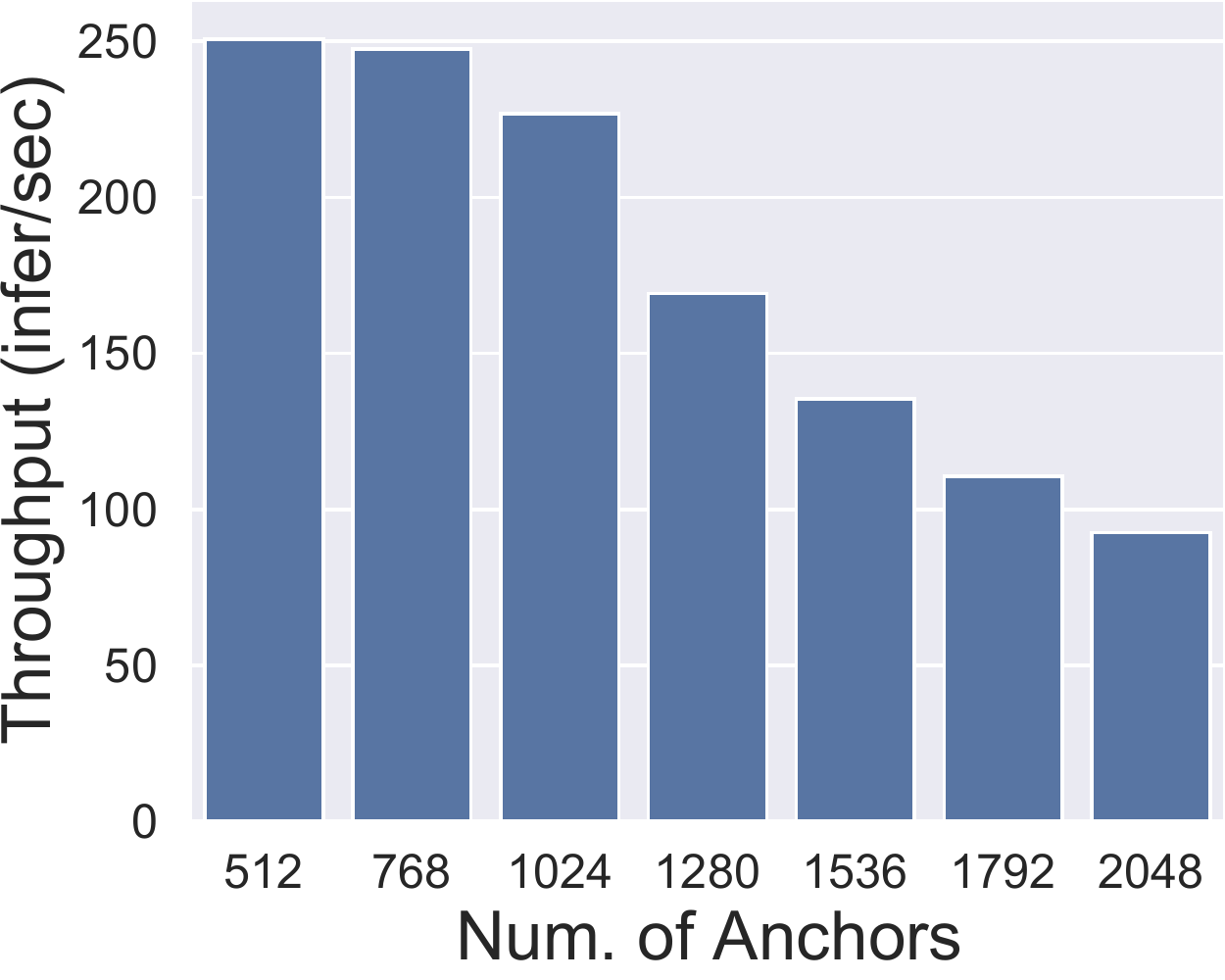}
        \caption{Inference throughput.
}
        \label{subfig:triton_throughput}
    \end{subfigure}
    \vspace{-1em}
    \caption{
        3D vision-based lighting estimator performance.
}
    \label{fig:xihenet}
    \vspace{-.5em}
\end{figure}

We evaluate the performance of \xiheNet and compare its accuracy to a state-of-the-art lighting estimator PointAR~\cite{pointar_eccv2020}.
We use the same 3D indoor dataset (about 608k training and 2037 test data) as PointAR and preprocess each image to generate the \uspc and \SHc pairs.
We extract \SHc from the LDR format given its best visual rendering effects. We repeat the same process using the uniform random sampling to generate the training dataset for PointAR.
We train both \xiheNet and PointAR using the same hyperparameters, i.e., 2 PointConv blocks, each with multilayer perceptron setup of (64, 128) and (128, 256)~\cite{pointar_eccv2020}.
Additionally, we train a third model (with the same backbone as \xiheNet) with farthest point sampling~\cite{pointnet_plus_plus_nips}.

Figure~\ref{fig:xihenet} shows the performance of the 3D vision-based estimators.
We use \SHc RMSE, which is defined as the numerical difference between the predicted and ground truth \SHc, as the metric to evaluate the lighting estimation accuracy~\cite{Garon2019}.
Our \xiheNet achieves better \SHc RMSE (the lower the better) for all \uspc sizes as shown in Figure~\ref{subfig:xihenet_accuracy}.
Further, we observe that \xiheNet on a Nvidia RTX 2080Ti GPU can generate lighting estimations between
3.99ms to 10.80ms. As shown in Figure~\ref{subfig:triton_throughput}, we can support up to 250 inferences per second with batch size=1.

\subsection{Triggering Strategy Analysis}

\begin{figure}[t]
    \centering
    \begin{subfigure}[b]{0.45\linewidth}
        \centering
        \includegraphics[width=\linewidth]{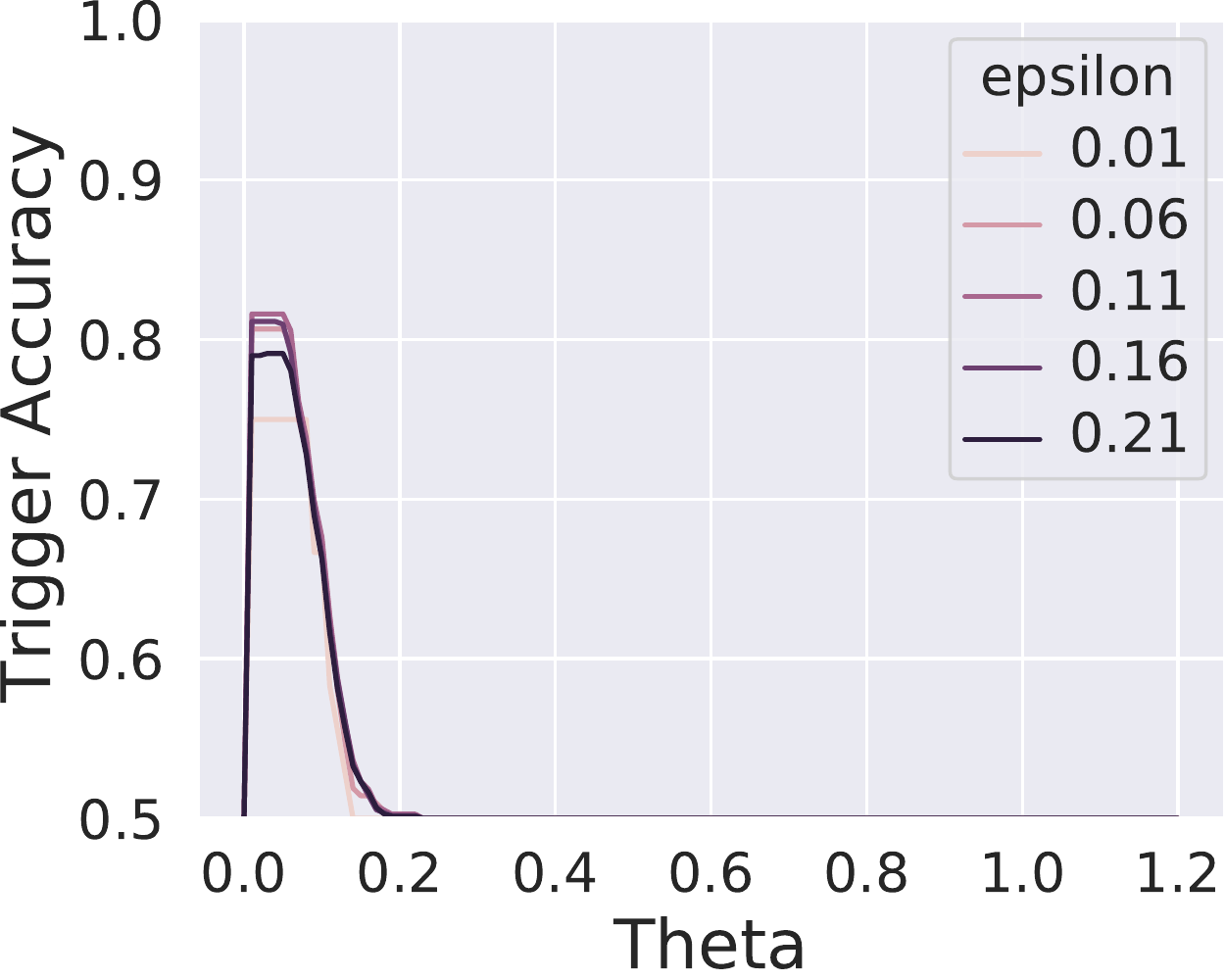}
        \vspace{-1.5em}
        \caption{Observation image MSE.}
    \end{subfigure}\quad
    \begin{subfigure}[b]{0.45\linewidth}
        \centering
        \includegraphics[width=\linewidth]{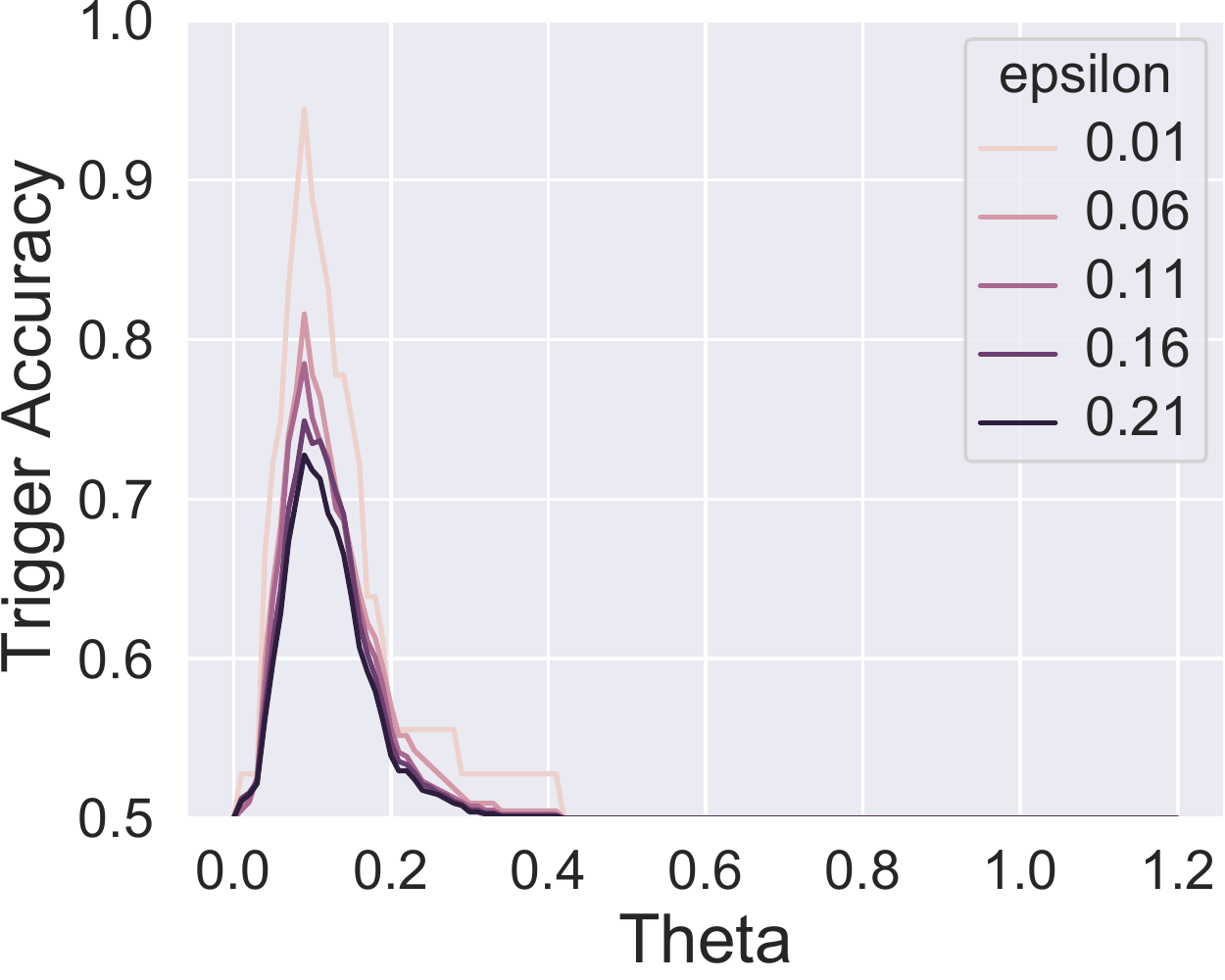}
        \vspace{-1.5em}
        \caption{\sysname window size=1.}
    \end{subfigure}
    \quad
    \begin{subfigure}[b]{0.45\linewidth}
        \centering
        \includegraphics[width=\linewidth]{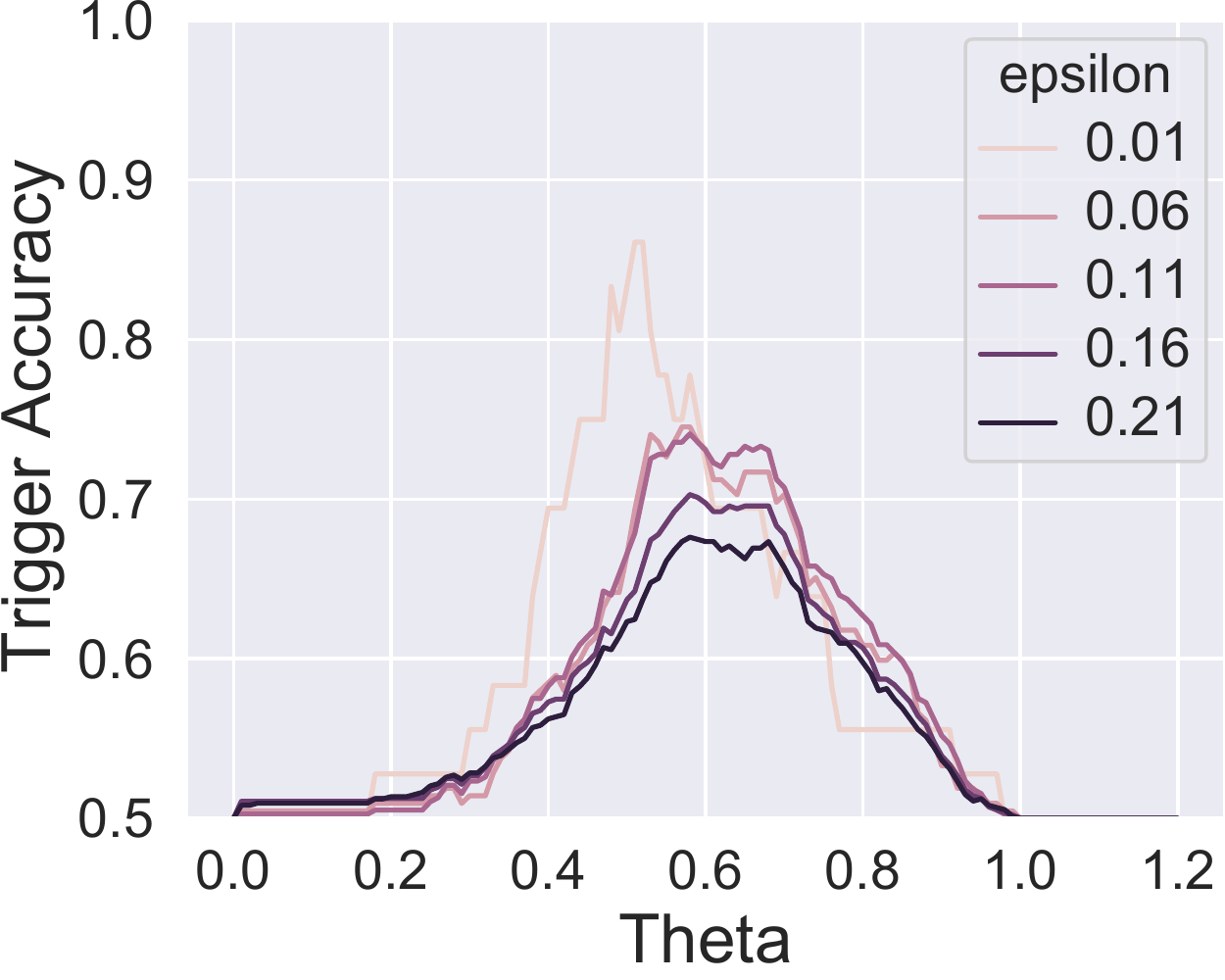}
        \vspace{-1.5em}
        \caption{\sysname window size=4.}
    \end{subfigure}
\quad
    \begin{subfigure}[b]{0.45\linewidth}
        \centering
        \includegraphics[width=\linewidth]{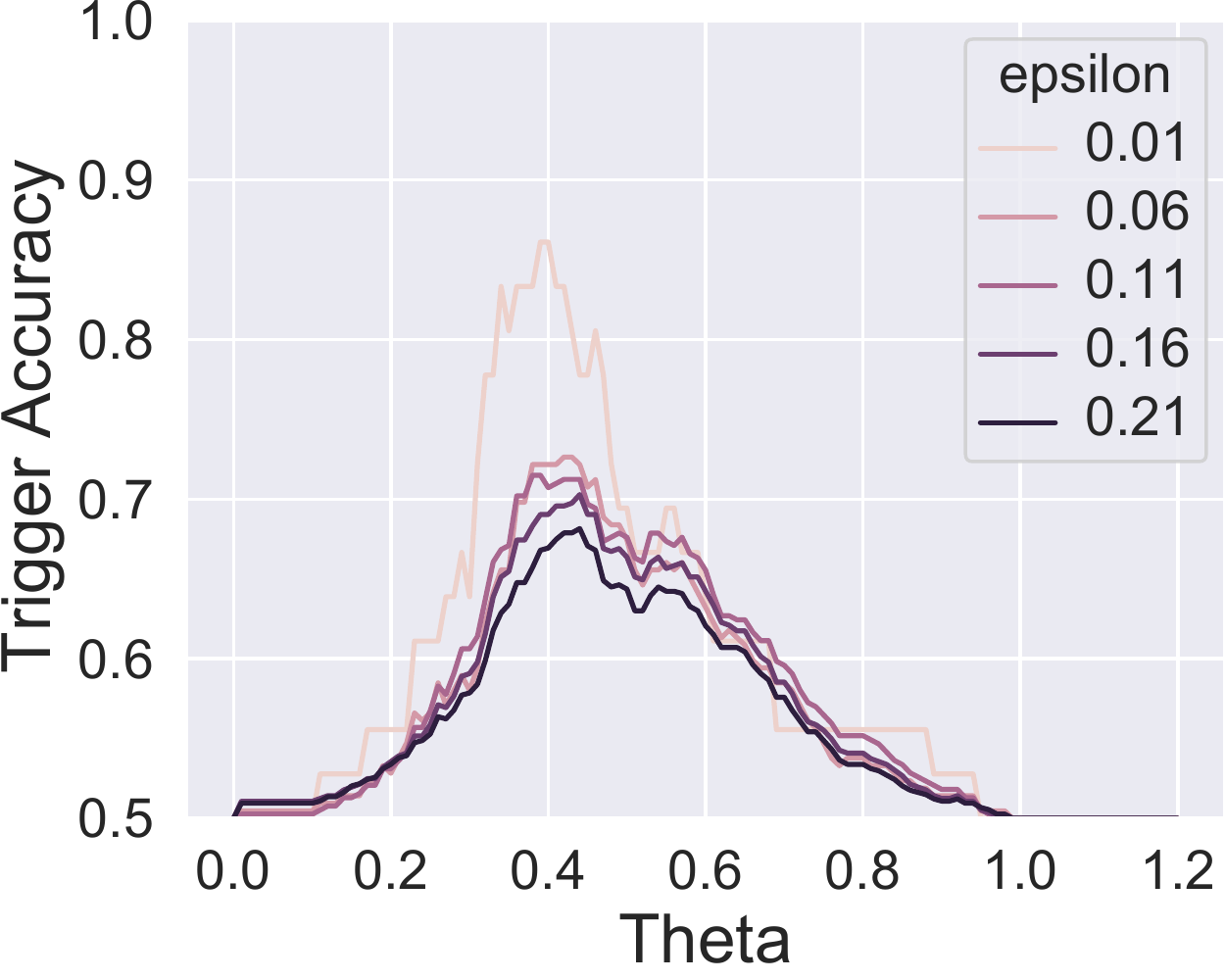}
        \vspace{-1.5em}
        \caption{\sysname window size=16.}
    \end{subfigure}
    \vspace{-1em}
    \caption{
        Triggering strategy analysis.
\textnormal{Each strategy is evaluated on five $\epsilon$-percentile split datasets.}
    }
    \label{fig:trigger_acc}
    \vspace{-1.5em}
\end{figure}

We quantify the effectiveness of different triggering strategies using the metric \emph{triggering accuracy} that describes the percentage of correctly identified environment lighting changes between two consecutive frames.
We create a new dataset that consists of 1754 image pairs and a binary label for each pair indicating the lighting change.
The image pairs are formed with two methods:
\1 select data items from the test dataset that share the same observation image but from different estimation locations;
\2 randomly select a pair of data items.The first pairing method covers the slower-changing scene scenarios while the second pairing method covers faster-changing ones.
To assign the lighting change label to each pair, we use three common metrics,
\SHc RMSE and two image-based metrics (reconstructed irradiance map PSNR and SSIM),
that are used for describing lighting conditions.
We then sort each image pair in ascending order based on the metric calculation and label the lower and higher $\epsilon$ percents as 0 and 1, respectively.
A label of 0 indicates no lighting change while a label of 1 indicates lighting change.
By ignoring the middle percentile, our label assignment method allows us to automatically distinguish the pairs that exhibit lighting changes from ones that do not with high confidence.
We manually verify the labeling results.

Figure~\ref{fig:trigger_acc} compares the triggering accuracy with different dataset partition threshold $\epsilon$ and triggering threshold $\theta$, using the SSIM metric.
Results corresponding to other metrics show similar trends and are omitted.
We first observe that the observation MSE-based triggering strategy has a good accuracy when the triggering threshold $\theta$ is properly configured.
Outside a small range of $\theta$, the triggering accuracy is significantly lower.
Second, we see that \sysname's sliding window based strategy can achieve better triggering accuracy than the MSE-based strategy and it's less sensitive to the choice of $\theta$.
Using larger window sizes have little impact on the triggering accuracy.
As different window sizes correspond to different computation complexity, we use a default of window size=4 in \sysname.

\setlength{\tabcolsep}{4pt}
\begin{table}[t]
\centering
\caption{
Real-world evaluation of \sysname.
\textnormal{We record and replay three AR sessions under different lighting and movement dynamics.}
}
\vspace{-0.5em}
\scriptsize{
\begin{tabular}{lrrrr}
\toprule
\textbf{Variable} & \textbf{Threshold $\theta$}
& \begin{tabular}[r]{@{}r@{}}\textbf{\SHc}\\ \textbf{RMSE Mean}\end{tabular}
& \begin{tabular}[r]{@{}r@{}}\textbf{\SHc}\\ \textbf{RMSE Std} \end{tabular}
& \begin{tabular}[r]{@{}r@{}}\textbf{Triggering}\\ \textbf{Percentage} \end{tabular} \\
\midrule
{} & 0.5 & 0.0180 & 0.0119 & 1.59\% \\
\textbf{R1:} Light temperature & \textbf{0.6} & \textbf{0.0193} & \textbf{0.0175} & \textbf{0.27\%} \\
{} & 0.7 & 0.0169 & 0.0047 & 0.13\% \\

\midrule

{} & 0.5 & 0.0113 & 0.0065 & 2.69\% \\
\textbf{R2:} Light intensity & \textbf{0.6} & \textbf{0.0110} & \textbf{0.0038} & \textbf{0.24\%} \\
{} & 0.7 & 0.0180 & 0.0054 & 0.24\% \\

\midrule

{} & 0.5 & 0.0040  & 0.0231 &  48.41\% \\
\textbf{R3:} User movement & \textbf{0.6} & \textbf{0.0070} & \textbf{0.0262} &  \textbf{23.67\%} \\
{} & 0.7 & 0.0102  & 0.0051 &  5.91\% \\

\bottomrule
\end{tabular}
}
\label{tab:recording}
\vspace{-1.5em}
\end{table}

\begin{figure*}
    \centering
\includegraphics[width=1\linewidth]{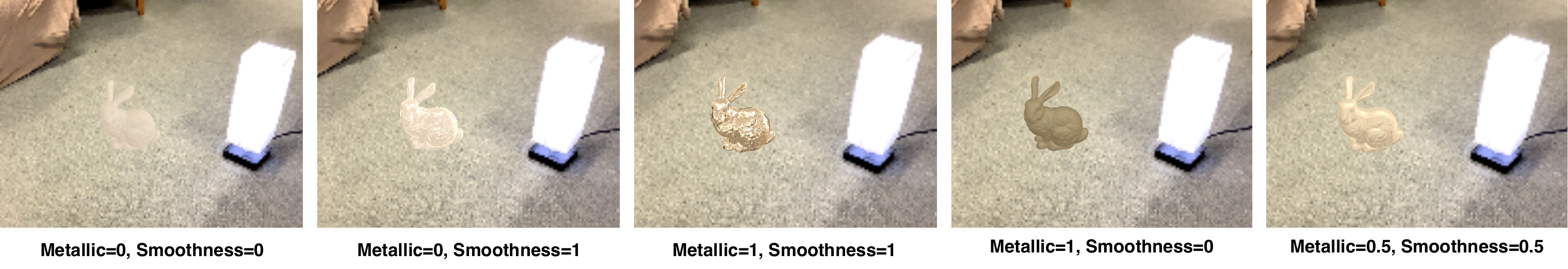}
\vspace{-2em}
    \caption{
AR scenes rendered with \sysname with different materials.
\textnormal{Both metallic and smoothness are Unity parameters.}
}
    \label{fig:eval_materials}
    \vspace{-1em}
\end{figure*}

\subsection{Lab-based Real-world Evaluation}

In this section, we present a real-world evaluation of \sysname in a lab environment to demonstrate its effectiveness of our triggering algorithm.
We show that with optimal configuration ($\theta=0.6$, $N=4$, 1280 anchors), \sysname can skip sending at least 76.24\% estimation requests to the edge while still achieves comparable accuracy to running inference every frame.
We use \sysname's session recorder to capture recordings when the user is interacting with our reference AR application in an indoor environment.
We create both lighting condition and movement dynamics by using a remotely controlled light source and having the user walk around the light source.

For each recording, we control one of the variables, i.e., light temperature, light intensity, and user movement.
The light source allows us to vary the temperature from candle light (1500K) to daylight (6500K) with 500K increment and the intensity from 0\% to 100\% (800 lumens) with 1\% increment.
We record relevant AR session information per frame. In total, we create three recordings with an average length of 35 seconds.
We replay each recorded AR session to \sysname and report both the \SHc RMSE and the percentage of triggered frames (i.e., being sent to the edge).

As shown in Table~\ref{tab:recording}, \sysname ($\theta=0.6$) only needs to send up to 23.67\% inference requests to the \xiheNet while only incurring an average RMSE of 0.011.
We also inspect the visual effects of the rendered object during the replay and confirm minimal differences with and without the triggering algorithm enabled.
Interestingly, for the third recording where the user is walking around the light source with the iPad Pro, \sysname triggers a lot more inference requests than the other two recordings.
We suspect that more frequent triggering is likely due to increased observation completeness at estimation positions and enlarged viewing angles.

\section{Related Work}
\label{sec:related}

To provide mobile AR that is suitable for real-world deployment, researchers have been working on aspects including energy optimization, interactivity, and lifelike rendering~\cite{Chen2018-xp,rohmer2017natural,Apicharttrisorn2019-fa}.
The key to achieve lifelike rendering in AR is the ability to obtain accurate lighting information~\cite{Garon2019,ramamoorthi_efficient_2001,srinivasan20lighthouse}. Although intuitively simple, there are a number of AR-specific challenges that distinguish this task from prior work in the  graphics community~\cite{debevec2006image,gunther2007realtime,ramamoorthi_efficient_2001}.

There has been little work on providing real-time lighting estimation for mobile AR~\cite{prakash2019gleam,pointar_eccv2020,ChengSCDZ18_graph}. Commercial SDKs such as ARKit~\cite{arkit} and ARCore~\cite{arcore} only provide ambient lighting estimation which is often insufficient to capture the spatially-variant environment lighting. A recent work GLEAM leverages physical probes and improves the rendering effects over these commercial SDKs~\cite{prakash2019gleam}. However, the use of physical probes hinders the user experiences.
Our work is the first 3D vision-based framework that provides spatially-variant lighting estimation in real time.

The system design of \sysname is largely inspired by our empirical study and tackles problems that are common to edge-based AR systems~\cite{Liu2018-we,Liu2019-pn,Liu2020-fy}. Specifically, when designing \sysname, we focus on minimizing the reliance on mobile resources to avoid excessive power consumption~\cite{Chen2018-xp,Apicharttrisorn2019-fa,Hu2019-rj}; we also minimize the network communication to the edge server by only issuing requests that are likely to improve the lighting estimation accuracy, e.g., when the lighting condition changes or when \sysname has more updated environment information~\cite{Chen2015-ce,Liu2018-aq}.
Our work differs from existing work on edge-based AR systems in addressing the lighting estimation-specific requirements and 3D vision-based opportunities.

\section{Discussion}
\label{sec:discussion}

We made the conscious decision to co-design some aspects of \sysname, including the \uspcs and the triggering metric, with a state-of-the-art 3D lighting estimator~\cite{pointar_eccv2020}.
We believe such application-specific optimizations are worthwhile trade-offs, allowing us to fully explore the performance potential of both algorithms and systems.
In other words, rather than exposing the trade-offs of accuracy and performance to AR developers, we offload such responsibilities to the framework design phase.

Nevertheless, the lighting estimation accuracy provided \sysname will be bounded by the supported lighting estimators. For example, as \xiheNet currently only demonstrates good estimation accuracy for low-frequency lighting, virtual objects that require high-frequency lighting information (such as metallic finish) will have less photorealistic rendering effects. Figure~\ref{fig:eval_materials} show the visual effects of Stanford bunny with different material settings.
Lower metallic values will give more matte-looking finish while lower smoothness values will lead to higher diffused reflection.

Additionally, to \emph{effectively} support any future models on \sysname, respective components have to be rethought and redesigned. However, given \sysname's modular design and that its major components are general enough, we do not anticipate substantial changes.
Lastly, \sysname currently provides an end-to-end 3D vision-based lighting estimation service per AR session. To support multi-user shared AR sessions, we will at least need to redesign the Point Cloud Management module to carefully manage the lifecycles and states.

\section{Conclusion}
\label{sec:conclusion}

Our system \sysname is a 3D vision-based lighting estimation platform that provides fast and accurate spatially-variant lighting estimation for mobile AR systems.
Specifically our \uspcs allows us to effectively downsample the raw point cloud captured in real time without compromising the lighting estimation accuracy.
To avoid unnecessary network communication between mobile and edge, we designed an adaptive triggering algorithm that only sends \uspc to the edge when there are significant lighting condition changes.
The good estimation accuracy is guaranteed by our 3D-based lighting model that is inspired by recent work~\cite{pointar_eccv2020,rohmer2017natural} and is redesigned to consider both network and storage cost.
We implemented \sysname on top of Unity3D, ARFoundation, and Pytorch frameworks.
Our controlled experiments with three devices including a Lidar-enabled iPad Pro demonstrated that \sysname can provide visually-better rendering than ARKit and GLEAM under various experiment settings.

\begin{acks}
We thank all anonymous reviewers, our shepherd, and our artifact evaluator Tianxing Li for their insight feedback.
This work was supported in part by NSF Grants \#1755659 and \#1815619.
\end{acks}

\balance
\bibliographystyle{ACM-Reference-Format}
\bibliography{main.bib}

\end{document}